\def\year{2020}\relax
\documentclass[letterpaper]{article} % DO NOT CHANGE THIS
\usepackage{aaai20}  % DO NOT CHANGE THIS
\usepackage{times}  % DO NOT CHANGE THIS
\usepackage{helvet} % DO NOT CHANGE THIS
\usepackage{courier}  % DO NOT CHANGE THIS
\usepackage[hyphens]{url}  % DO NOT CHANGE THIS
\usepackage{graphicx} % DO NOT CHANGE THIS
\usepackage{graphicx}  % DO NOT CHANGE THIS
\usepackage{booktabs}
\usepackage{amssymb}
\usepackage{amsmath}
\usepackage{subcaption}
\usepackage{algorithm}
\usepackage{algorithmic}
\usepackage{hyperref}

\newcommand{\citet}[1]{\citeauthor{#1} \shortcite{#1}}
\newcommand{\citep}{\cite}
\newcommand{\Algref}[1]{Algorithm~\ref{#1}}

\nocopyright

\title{An Adaptive and Momental Bound Method for Stochastic Learning}
\author{
Jianbang Ding, Xuancheng Ren, Ruixuan Luo, Xu Sun \\
MOE Key Laboratory of Computational Linguistics, School of Electronics Engineering and Computer Science, Peking University \\
\texttt{\{jianbangding, renxc, luoruixuan97, xusun\}@pku.edu.cn}
}

\begin{document}

\maketitle

\begin{abstract}
Training deep neural networks requires intricate initialization and careful selection of learning rates. The emergence of stochastic gradient optimization methods that use adaptive learning rates based on squared past gradients, e.g., AdaGrad, AdaDelta, and Adam, eases the job slightly. However, such methods have also been proven problematic in recent studies with their own pitfalls including non-convergence issues and so on. Alternative variants have been proposed for enhancement, such as AMSGrad, AdaShift and AdaBound. In this work, we identify a new problem of adaptive learning rate methods that exhibits at the beginning of learning where Adam produces extremely large learning rates that inhibit the start of learning. We propose the \textbf{Ada}ptive and \textbf{Mo}mental Boun\textbf{d} (AdaMod) method \footnote{Our implementation is available at: \\ \href{https://github.com/lancopku/AdaMod}{\texttt{https://github.com/lancopku/AdaMod}}.} to restrict the adaptive learning rates with adaptive and momental upper bounds. The dynamic learning rate bounds are based on the exponential moving averages of the adaptive learning rates themselves, which smooth out unexpected large learning rates and stabilize the training of deep neural networks. Our experiments verify that AdaMod eliminates the extremely large learning rates throughout the training and brings significant improvements especially on complex networks such as DenseNet and Transformer, compared to Adam.
\end{abstract}

\section{Introduction}
Gradient-based optimization forms the core of first-order optimization algorithms to train deep networks today. Remarkably, stochastic gradient descent (SGD) \cite{robbins1951stochastic}, one of the most dominant methods, performs well across many applications, despite its simplicity.
However, one shortcoming of SGD is that it scales the gradient uniformly in all directions. This strategy requires a subtle tuning of the learning rate and limits the training speed in the early stage. To address this issue, several adaptive methods have been proposed to achieve faster convergence by computing individual learning rates for different parameters. Examples of such methods include AdaGrad~\cite{duchi2011adaptive}, Adam~\cite{DBLP:journals/corr/KingmaB14}, RMSProp~\cite{tieleman2012lecture} and AdaDelta~\cite{zeiler2012adadelta}. They use adaptive moment estimation of the past squared gradients to adjust the individual learning rates. In particular, Adam is regarded as the default algorithm used across many deep learning frameworks \cite{wilson2017marginal}.

Although adaptive methods gain great popularity in many settings, they still stumble on the stability problem. \citet{DBLP:conf/iclr/ReddiKK18} focused on the non-convergence issue of Adam, and pointed out the lack of ``long-term memory'' in Adam-like algorithms, which hamper their performance and lead to divergence. Recently, \citet{DBLP:conf/iclr/LuoXLS19} proposed a variant of Adam called AdaBound to solve this problem. The authors illustrated that the lack of generalization performance of adaptive methods may stem from unstable and extreme learning rates, and proposed to clip the extreme learning rates by employing dynamic bounds on them. However, AdaBound only dealt with the extreme learning rates at the end of training, and ignored those in the early stage, which may also cause training instability and lead to divergence, especially for complex neural networks. 

\begin{figure*}[t]
    %\centering

    \subcaptionbox{Training loss  \label{fig:contrast}}{\includegraphics[height=10\baselineskip]{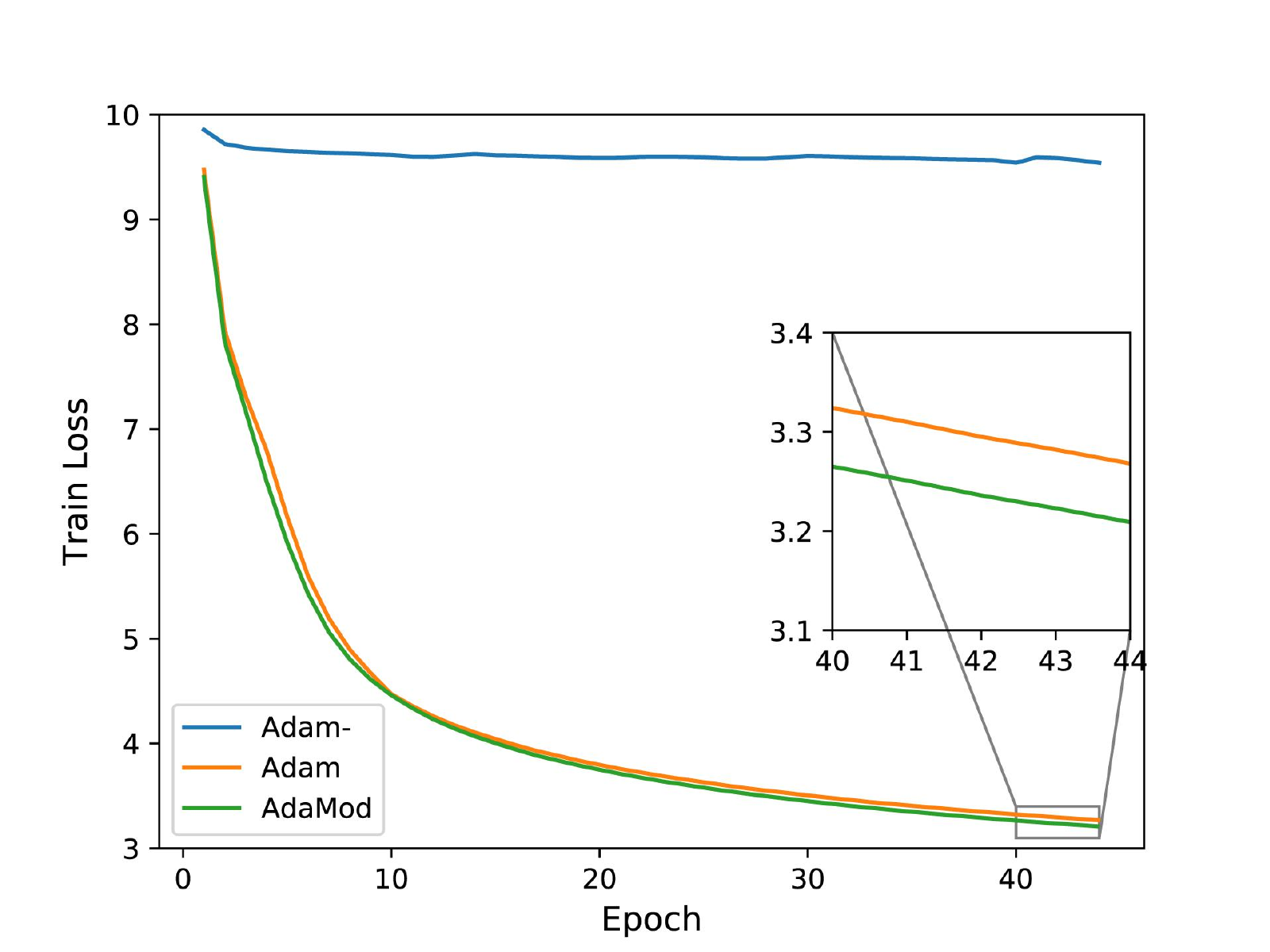}}
    \hfil
    \subcaptionbox{Learning rates for Adam without warmup \label{fig:d0}}[0.35\linewidth]{\includegraphics[height=10\baselineskip]{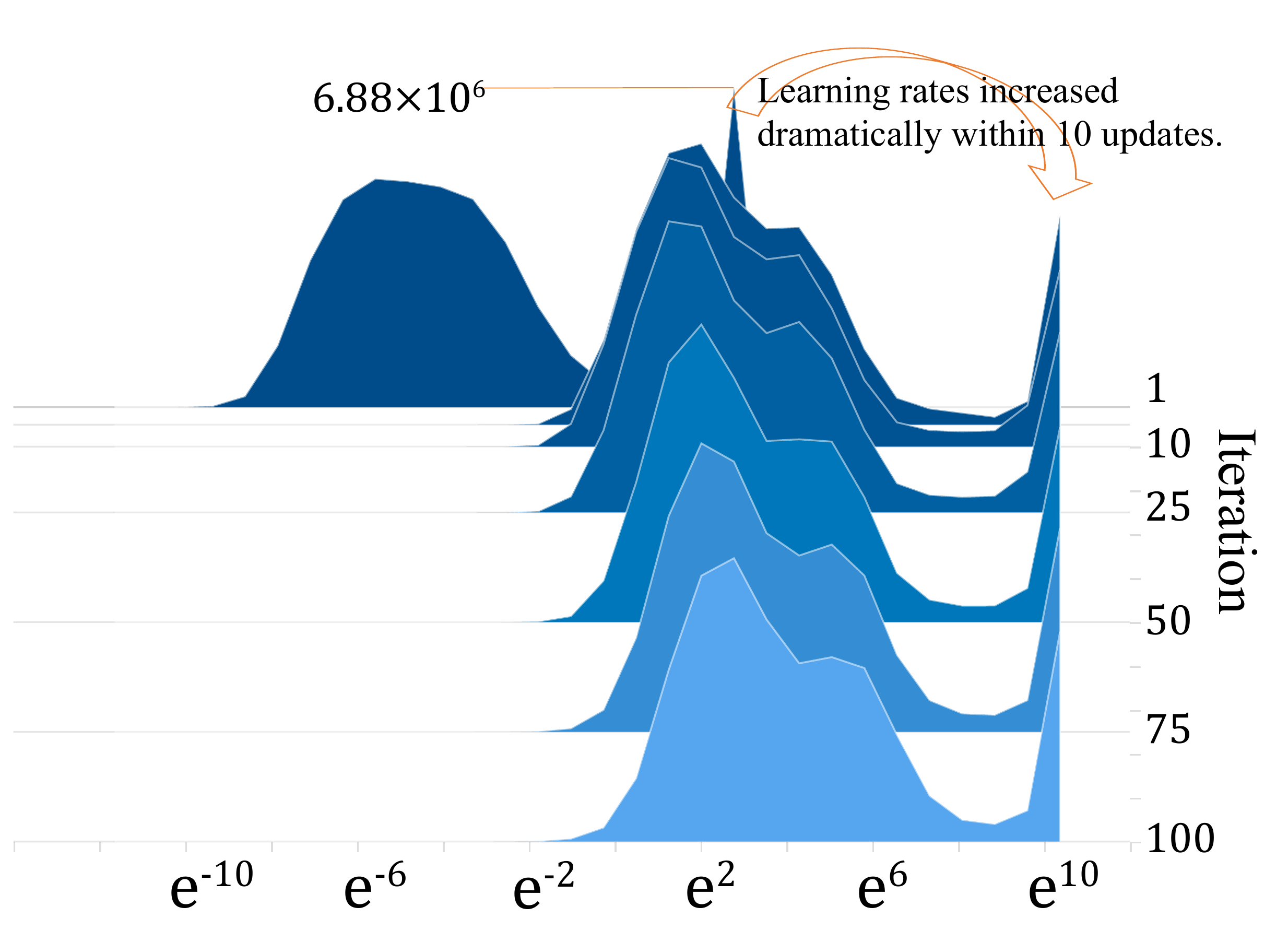}}
    \hfil
    \subcaptionbox{Learning rates for Adam with warmup    \label{fig:d1}}{\includegraphics[height=10\baselineskip]{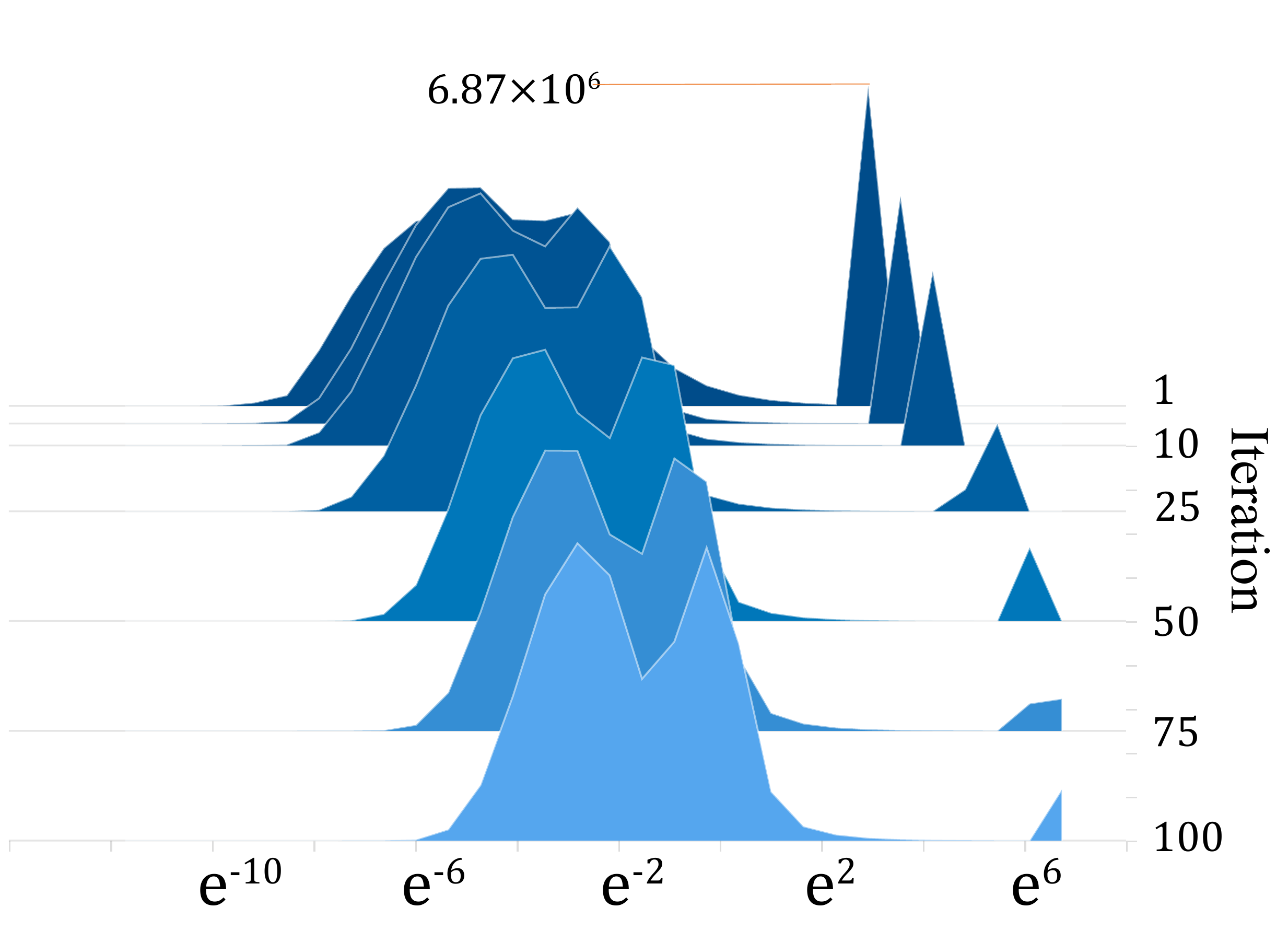}}

    \caption{Training loss and learning rate distribution of Transformers on the IWSLT'14 De-En dataset. ``Adam-'' in  (a) denotes Adam without warmup. For (b) and (c), X-axis is original value in the log scale; Y-axis is training iterations and the height stands for frequency. Adam does not converge without warmup due to extremely large learning rates, while AdaMod can fix this issue and perform better.}
    \label{fig:observation}
\end{figure*}

%As mentioned above, we have identified that adaptive methods sometimes appear to be unstable during early training stage as extreme learning rates, especially on complex deep networks. 
Learning rate warmup scheme is hence motivated as a common heuristic to train complex neural networks without causing instability by starting with small learning rates and increasing them gradually in the first few epochs \cite{DBLP:conf/iclr/GotmareKXS19}. For example, on the IWSLT'14 De-En dataset, removing warmup assistance could result in a sharp increase of learning rates in the first 10 updates, meanwhile the training loss fluctuates around 9.5 and hardly decreases, as shown in Figure \ref{fig:observation}. Similar phenomena are observed in other tasks such as Transformer-XL \cite{DBLP:conf/acl/DaiYYCLS19} language modeling. In the absence of theoretical guarantees of the warmup heuristic, researchers usually need to experiment with different hyperparameter settings across different networks or tasks, which consumes a lot of time.

In this paper, we first conduct an empirical study on the warmup heuristic and illustrate that the the great variance of the adaptive learning rates in the early training stage can account for the extremely large rates. These may increase the probability of oscillating between local optima, causing non-convergence problems, and leading to poor generalization performance, which hardly raise concerns of most optimization algorithms.

%%%%%%%%%%%%%%%%%%%%%%%%%%%%%%%%%%%%%%%%%%%%%%%%%%%%%%%%%%%%%%%%%%%%%%%%%%%%%%%%%%%%%
%%%%%%%%%%%%%%%%%%%%%%%%%%%%%%%%%%%%%%%%%%%%%%%%%%%%%%%%%%%%%%%%%%%%%%%%%%%%%%%%%%%%%

Under this premise, we propose a new variant of Adam, AdaMod, to restrict the adaptive learning rates with adaptive and momental upper bounds. We aim to smooth out unexpected large learning rates and stabilize the training process based on the adaptive learning rates themselves. Specifically, we apply exponential moving averaging to the adaptive learning rates computed by Adam to get the smoothed learning rates, and then employ them as an upper bound on the original. This endows learning rates with ``long-term-memory'' of past gradients in order to improve their stability. With this framework, we can obtain a stable training of good generalization performance, and reduce training hyperparameters in many settings (e.g. get rid of the warmup scheme).

Finally, we conduct further experiments on various models and tasks in computer vision and natural language processing. Empirical results demonstrate that our method can effectively avoid unexpected large learning rates in the training process and can hence fix the non-convergence problem. Moreover, it can bring considerable improvement over the vanilla Adam especially on complex deep networks. 

\section{Background}

\subsubsection{A Brief Review of Adam}

%%%%%%%%%%%%%%%%%%%%%%%%%%%%%%%%%%%%%%%%%%%%%%%%%%%%%%%%%%%%%%
\begin{algorithm}[t]
	\caption{Adam}
	\label{alg:adam}
	\begin{algorithmic}[1]
		\REQUIRE initial parameter $\theta_{0}$, step sizes $\{\alpha_{t}\}_{t=1}^{T}$, moment decay $\{\beta_{1},\beta_{2}\}$, regularization constant $\epsilon$, stochastic objective function $f(\theta)$
		\STATE Initialize $m_{0}=0$, $v_{0}=0$
		\FOR{$t=1$ \TO $T$}
		\STATE $g_{t}=\nabla  f_{t}(\theta_{t-1})$
		\STATE $m_{t}=\beta_{1}m_{t-1}+(1-\beta_{1})g_{t}$
		\STATE $v_{t}=\beta_{2}v_{t-1}+(1-\beta_{2})g_{t}^{2}$
		\STATE $\hat{m}_{t}=m_{t}/(1-\beta_{1}^{t})$
		\STATE $\hat{v}_{t}=v_{t}/(1-\beta_{2}^{t})$
		\STATE $\eta_{t}=\alpha_{t}/(\sqrt{\hat{v}_{t}}+\epsilon)$
		\STATE $\theta_{t}=\theta_{t-1}-\eta_{t}\hat{m}_{t}$
		\ENDFOR
	\end{algorithmic}
\end{algorithm}
%%%%%%%%%%%%%%%%%%%%%%%%%%%%%%%%%%%%%%%%%%%%%%%%%%%%%%%%%%%%%%

\Algref{alg:adam} provides a brief review of Adam for reference. The setup is elaborated as follows. We first compute the gradient $g_{t}$ of the loss function with respect to previous parameters. Second, we update the low-order moments of gradient $m_{t}$, $v_{t}$ by adopting exponential averaging and compute bias-corrected versions for them. Finally, we refresh the parameter to get a new $\theta_{t}$. This process needs to iterate $T$ steps until we return our learned parameters.

\subsubsection{Warmup learning rate scheme}
Generally, a simple constant step size $\alpha_{t}$ as well as a decreasing scheme on it both work well in practice. But in some cases, researchers have to adopt a step size increasing strategy in the early training stage such as the warmup scheme. Specifically, some extra hyperparameters have to be set including a small step size initial value $\alpha_{0}$, a step size target value $\alpha_{w}$, update steps of warmup $T_{w}$ and rules for step size growth (e.g. linear growth sets $\alpha_{t}=\alpha_{0}+\frac{\alpha_{w}-\alpha_{0}}{T_{w}}t$, when $t<T_{w}$). Warmup is regarded as a means to use large learning rates and avoid non-convergence problems. Although it lacks strong theoretical support, it has been beneficial in many deep learning tasks. 

\subsubsection{Extremely large learning rates leading to instability issues}

% %%%%%%%%%%%%%%%%%%%%%%%%%%%%%%%%%%%%%%%%%%%%%%%%%%%%%%%%%%%%%%%%%%%%%%%%%%%%%%%%%%%%%
% \begin{figure}[t]
%     \centering
%     \includegraphics[width=0.4\textwidth]{img/IWSLT/lr_distribution/contrast.png}
%     \caption{Training loss of Transformers on the IWSLT'14 De-En.}
%     \label{fig:contrast}
% \end{figure}
% %%%%%%%%%%%%%%%%%%%%%%%%%%%%%%%%%%%%%%%%%%%%%%%%%%%%%%%%%%%%%%%%%%%%%%%%%%%%%%%%%%%%%

% %%%%%%%%%%%%%%%%%%%%%%%%%%%%%%%%%%%%%%%%%%%%%%%%%%%%%%%%%%%%%%%%%%%%%%%%%%%%%%%%%%%%%
% \begin{figure}[t]
%     \hfil
%     \subcaptionbox{Adam without warmup   \label{fig:d0}}{\includegraphics[width=0.4\textwidth]{img/IWSLT/lr_distribution/adam_without_warmup_d0.png}}
%     \hfil
%     \subcaptionbox{Adam with warmup    \label{fig:d1}}{\includegraphics[width=0.4\textwidth]{img/IWSLT/lr_distribution/adam_withwarmup_d0.png}}
%     \hfil
    
%     \caption{The learning rate histogram of Transformers on the IWSLT'14 De-En.}
%     \label{fig:lr-distribution}
% \end{figure}
% %%%%%%%%%%%%%%%%%%%%%%%%%%%%%%%%%%%%%%%%%%%%%%%%%%%%%%%%%%%%%%%%%%%%%%%%%%%%%%%%%%%%%
Exploring how to tackle the non-convergence issue of adaptive methods is an important research interest of current machine learning research. In recent years, many remarkable works have provided us with better understanding of this problem with the proposal of different variants of Adam. \citet{DBLP:conf/iclr/ReddiKK18} first indicated that Adam may not converge due to the lack of ``long-term-memory'' of past gradients and provided a theoretical guarantee of convergence. Following this track, most of the previous studies focused on how to modify the re-scaling term $v_{t}$. \citet{DBLP:conf/iclr/ZhouZLWZY19} argued that there exists an inappropriate correlation between $g_{t}$ and $v_{t}$, which may result in unbalanced updates of step size. Therefore, the authors proposed to decorrelate them by temporal shifting, i.e. replacing $g_{t}$ with $g_{t-n}$ for some manually chosen $n$ to calculate $v_{t}$. In a similar vein, \citet{DBLP:conf/ijcai/HuangWD19} discussed that the past gradients $\{g_{1},...,g_{t-1}\}$ are more reliable than $g_{t}$. And the authors proposed to weight more of the all past gradients when designing $v_{t}$. However, these methods do not radically avoid the non-convergence problem in practice due to the existence of unexpected large learning rates.

To solve this problem, \citet{DBLP:conf/icml/ShazeerS18} considered to drop momentum and remove the larger-than-desired updates by selecting a threshold $d$ for update clipping. However, as their main goal is to minimize the memory cost of optimization algorithms, this technique remains less explored and has a limited improvement on generalization performance. To this end, \citet{DBLP:conf/iclr/LuoXLS19} implemented a gradual transition from Adam to SGD by employing dynamic bounds on learning rates to avoid extremely larger ones. However, its bound function is manually designed and the performance rely heavily on the selection of the final learning rate $\alpha^*$ of SGD.

As mentioned in Adabound \citep{DBLP:conf/iclr/LuoXLS19} , unstable and extreme learning rates usually appear at the end of training, which jeopardizes the generalization performance of adaptive methods. However, we further investigate that early-stage extreme learning rates, not only those at the end, can also worsen generalization performance and even lead to non-convergence problem. For example, in the NMT experiment in Figure \ref{fig:contrast} , the training loss converges to around 9.5 without warmup heuristic, and it decreases to below 3.5 after using warmup. In addition, the learning rate histogram are shown in Figure \ref{fig:d0} and Figure~\ref{fig:d1}, where the X-axis is original value in the log scale, Y-axis is iteration steps and the height stands for frequency. We can observe that without using the warmup scheme, there are lots of learning rates soaring over 10,000 compared to using it. Such extremely large learning rates may lead to oscillation of the sequence and trap the adaptive method in a exceptionally bad local optima. Meanwhile they can not help the optimization escape from that, resulting in a series of non-convergence problems. These phenomena confirm our views above.

Despite all the previous efforts, the training stability of the Adam-like algorithms still waits for improvement, especially on complex networks. In this paper, we investigate the non-convergence issue from training Transformer-based model by Adam without warmup scheme, and this allows us to better understand the negative impact of extremely large learning rates and resolve the problem with a more concise and effective method.

\section{Methods} 
%%%%%%%%%%%%%%%%%%%%%%%%%%%%%%%%%%%%%%%%%%%%%%%%%%%%%%%%%%%%%%
\begin{algorithm}[t]
	\caption{AdaMod}
	\label{alg:AdaMod}
	\begin{algorithmic}[1]
		\REQUIRE initial parameter $\theta_{0}$, step sizes $\{\alpha_{t}\}_{t=1}^{T}$, moment decay $\{\beta_{1},\beta_{2}, \beta_{3}\}$, regularization constant~$\epsilon$, stochastic objective function $f(\theta_{0})$
		\STATE Initialize $m_{0}=0$, $v_{0}=0$, $s_{0}=0$
		\FOR{$t=1$ \TO $T$}
		\STATE $g_{t}=\nabla  f_{t}(\theta_{t-1})$
		\STATE $m_{t}=\beta_{1}m_{t-1}+(1-\beta_{1})g_{t}$
		\STATE $v_{t}=\beta_{2}v_{t-1}+(1-\beta_{2})g_{t}^{2}$
		\STATE $\hat{m}_{t}=m_{t}/(1-\beta_{1}^{t})$
		\STATE $\hat{v}_{t}=v_{t}/(1-\beta_{2}^{t})$
		\STATE $\eta_{t}=\alpha_{t}/(\sqrt{\hat{v}_{t}}+\epsilon)$
		\STATE $s_{t}=\beta_{3}s_{t-1}+(1-\beta_{3})\eta_{t}$
		\STATE $\hat{\eta}_{t}=\min(\eta_{t},s_{t})$
		\STATE $\theta_{t}=\theta_{t-1}-\hat{\eta}_{t}\hat{m}_{t}$
		\ENDFOR
	\end{algorithmic}
\end{algorithm}
%%%%%%%%%%%%%%%%%%%%%%%%%%%%%%%%%%%%%%%%%%%%%%%%%%%%%%%%%%%%%%

This section describes the AdaMod method as well as its properties, with the aim of reducing learning rates during the whole training process. concisely, AdaMod casts dynamic upper bounds on the adaptive learning rates that prevent the calculated learning rates from escalating too fast and becoming undesirably larger than what the historical statistics suggest. This helps control the variance of the adaptive learning rates and smooths out the out-of-expect fluctuations in the adaptive learning rates. The name AdaMod springs from \textbf{Ada}ptive and \textbf{Mo}mental Boun\textbf{d}. Pseudocode is provided in Algorithm \ref{alg:AdaMod}. 

\subsubsection{Smoothing adaptive learning rates}
Based on Adam, which computes adaptive learning rates with estimates of first and second moments (i.e. mean and uncentered variance) of the gradients, our method further estimates the first order moments of the individual adaptive learning rates $\eta_{t}$. Inspired by exponential moving average (EMA) which enjoys popularity in estimating the lower-order moments of the gradients. We do averaging directly on the learning rates $\eta_{t}$ computed by Adam. Specifically, we apply the following operation in Adam:
\begin{equation}
    s_{t}=\beta_{3}s_{t-1}+(1-\beta_{3})\eta_{t},
    \label{formula:f1}
\end{equation}
where $\eta_{t}$ are the learning rates computed by Adam at step $t$. Thus, the current smoothed value $s_{t}$ is an interpolation between the previous smoothed value $s_{t-1}$ and the current learning rates. The new hyperparameter $\beta_{3}$ controls the smoothness of $s_t$, as the average range of the data in the exponential moving average is $1/\beta_{3}$ (By evaluating its expansion form according to $t$). For example, when $\beta_3=0.9$ the average range is 10 periods; when $\beta_3=0.999$ the average range is 1,000 periods, so on and so forth. It is worth noting that when $\beta_3\rightarrow 0$, AdaMod is exactly equivalent to Adam.

Equation \ref{formula:f1} can be expressed in another version, where the current smoothed value is an exponentially weighted moving average with discount factor $\beta_{3}$:
\begin{equation}
    s_{t}=(1-\beta_{3})[s_{t-1}+\beta_{3}s_{t-2}+\beta_{3}^{2}s_{t-3}+...+\beta_{3}^{t-1}s_{0}].
    \label{formula:f2}
\end{equation}
This endows the current value $s_{t}$ with ``long-term-memory'' of past values $\{s_{t-1},...,s_{0}\}$. In practice, we set $s_{0}=0$ and do not apply bias correction to it in our method.
\subsubsection{Bounding adaptive learning rates}
For the current smoothed value $s_{t}$, we further take it as an adaptive upper bound for $\eta_{t}$ to eliminate extremely learning rates.
\begin{equation}
    \hat{\eta}_{t}=\min(\eta_{t},s_{t}),
    \label{formula:f3}
\end{equation}
where $\hat{\eta}_{t}$ is the final learning rates obtained by the bounding operation. Intuitively, this operation can be seen as clipping the learning rates element-wisely so that the output is constrained by the current smoothed value. Then we use $\hat{\eta}_{t}$ and $m_{t}$ to make a parameter update. This process needs to iterate $T$ steps until an approximate solution is returned.

%%%%%%%%%%%%%%%%%%%%%%%%%%%%%%%%%%%%%%%%%%%%%%%%%%%%%%%%%%%%%%%%%%%%%%%%%%%%%%%%%%%%%
\begin{table}[t]
    \centering
    \small
    \begin{tabular}{@{}l l l@{}}
    \toprule
    \textbf{Dataset} & \textbf{Network Type} & \textbf{Architecture} \\
    \midrule
    CIFAR-10     &  Deep Conv & ResNet-34 \\
    CIFAR-10     &  Deep Conv & DenseNet-121 \\
    CIFAR-100     &  Deep Conv & ResNet-34 \\
    CIFAR-100     &  Deep Conv & DenseNet-121 \\
    Penn Treebank & Recurrent & 3-Layer LSTM \\
    IWSLT'14 De-En & Attention & Transformer-Small \\
    WMT'14 En-De & Attention & Transformer-Base \\
    WMT'14 En-De & Attention & Transformer-Big \\
    \bottomrule
    \end{tabular}
    
    \caption{Details of the models for experiments.} 
    \label{tab:tab1}
\end{table}
%%%%%%%%%%%%%%%%%%%%%%%%%%%%%%%%%%%%%%%%%%%%%%%%%%%%%%%%%%%%%%%%%%%%%%%%%%%%%%%%%%%%%

%%%%%%%%%%%%%%%%%%%%%%%%%%%%%%%%%%%%%%%%%%%%%%%%%%%%%%%%%%%%%%%%%%%%%%%%%%%%%%%%%%%%%
\begin{figure*}[t]
    \hfil
    \subcaptionbox{Transformer-Small    \label{fig:loss1}}{\includegraphics[width=0.3\textwidth]{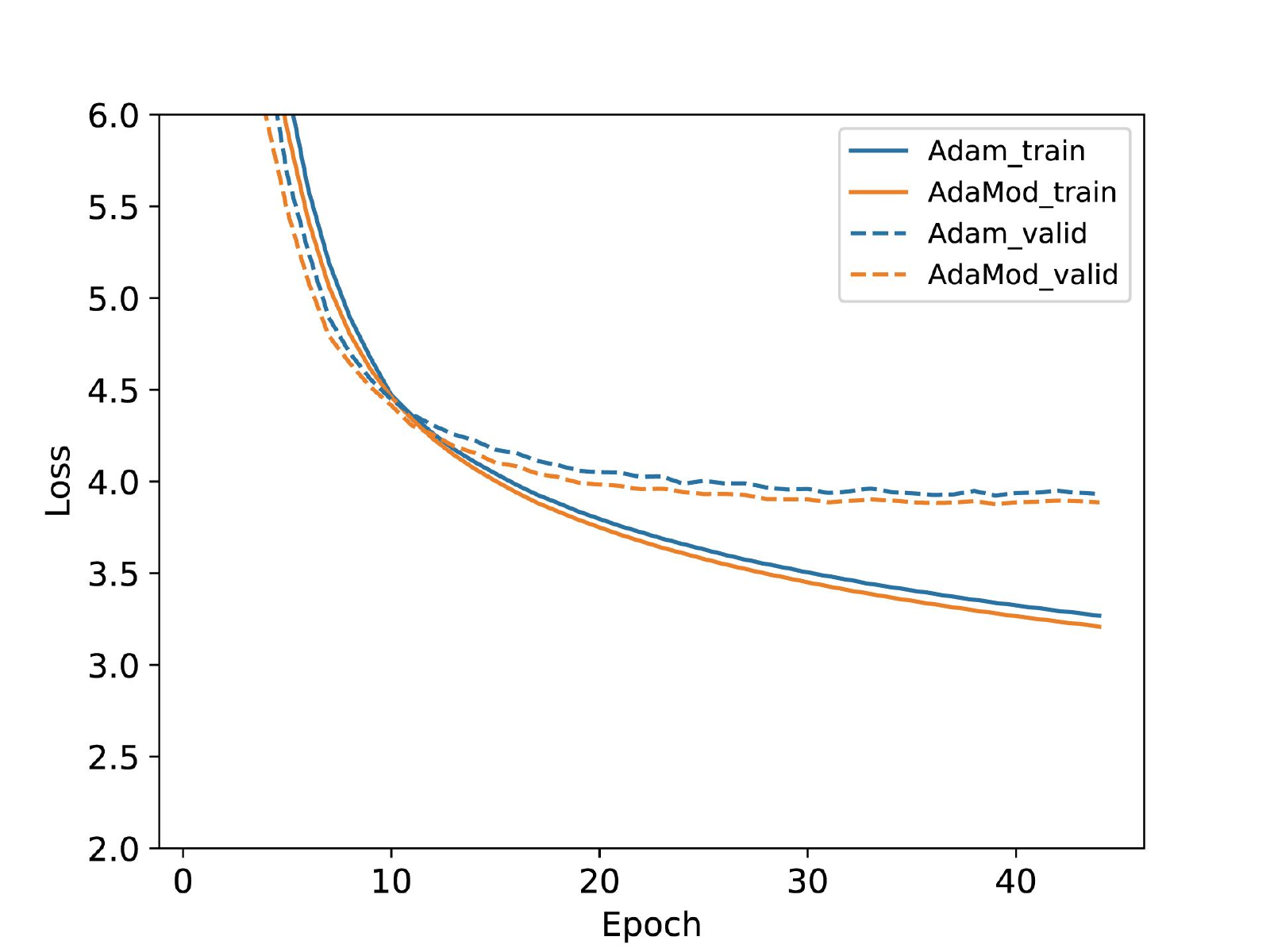}}
    \hfil
    \subcaptionbox{Transformer-Base    \label{fig:loss2}}{\includegraphics[width=0.3\textwidth]{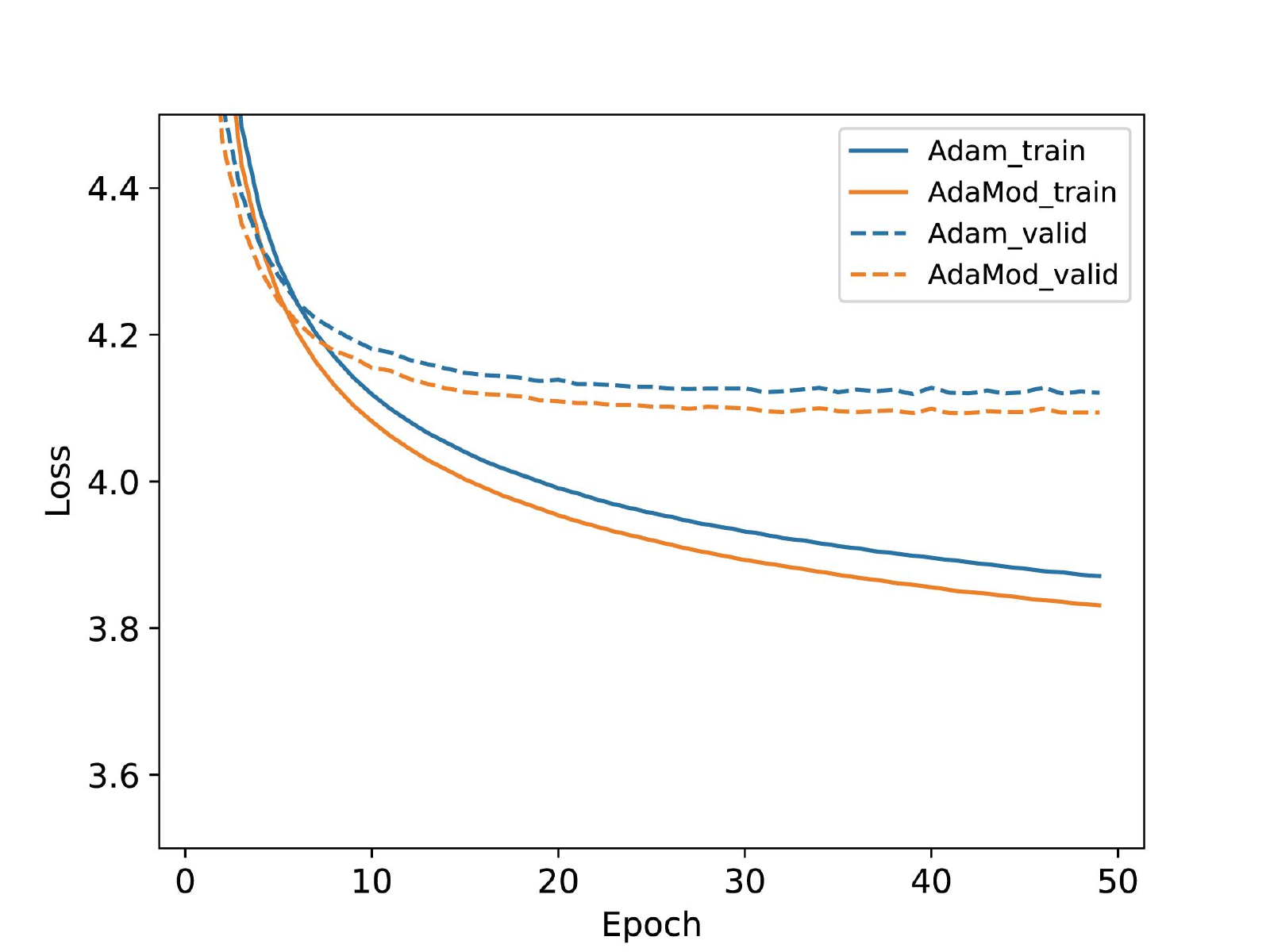}}
    \hfil
    \subcaptionbox{Transformer-Big    \label{fig:loss3}}{\includegraphics[width=0.3\textwidth]{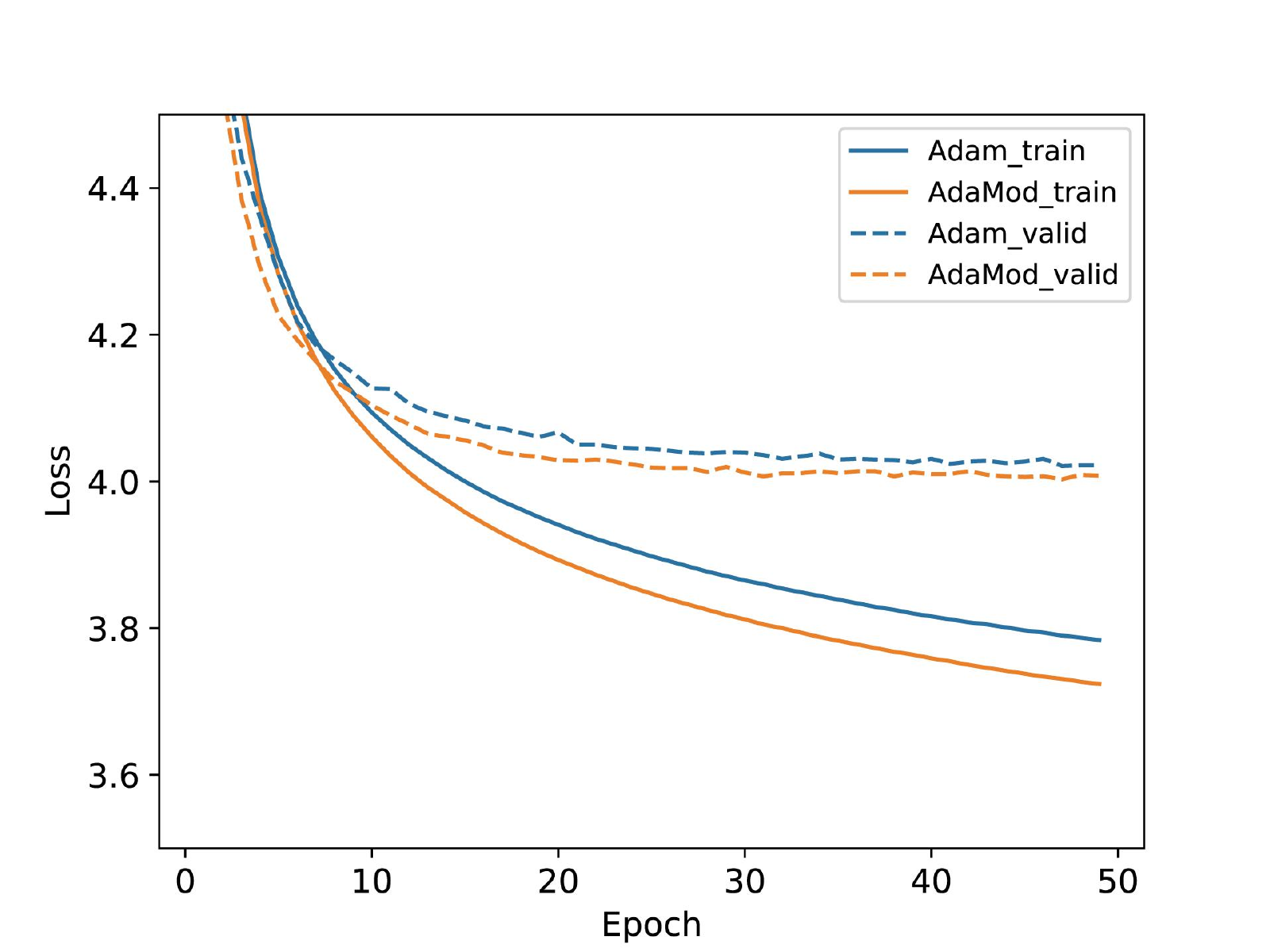}}
    \hfil
    
    \caption{Training and valid loss for Transformer-based model. For (a) is trained on IWSLT'14 De-En, (b) and (c) on WMT'14 En-De. AdaMod  without warmup shows both faster convergence and strong performance compared to Adam with warmup.}
    \label{fig:transformer-loss}
\end{figure*}
%%%%%%%%%%%%%%%%%%%%%%%%%%%%%%%%%%%%%%%%%%%%%%%%%%%%%%%%%%%%%%%%%%%%%%%%%%%%%%%%%%%%%

\section{Experiments}

This section performs a thorough evaluation of AdaMod optimizer on different deep learning tasks against fine-tuned baselines. We refer to several benchmarks: image classification on CIFAR-10/CIFAR100 \citep{krizhevsky2009learning}, language modeling on Penn Treebank \citep{marcus1993building}, and IWSLT'14 De-En/WMT'14 En-De for neural machine translation. The setup for each task is described in Table \ref{tab:tab1}. To achieve better performance, we apply decoupled weight decay to all adaptive methods in our experiment, on the basis of \citet{loshchilov2017fixing}'s work.

%%%%%%%%%%%%%%%%%%%%%%%%%%%%%%%%%%%%%%%%%%%%%%%%%%%%%%%%%%%%%%%%%%%%%%%%%%%%%%%%%%%%%
\begin{table}[t]
    \centering
    \small
    \begin{tabular}{@{}l c c@{}}
    \toprule
    \bf IWSLT'14 De-En  & \multicolumn{2}{c}{Transformer-Small}                   \\\midrule
    Adam without warmup & \multicolumn{2}{c}{/}          \\
    Adam with warmup   & \multicolumn{2}{c}{34.62}                         \\
    AdaMod              & \multicolumn{2}{c}{34.81}                         \\\midrule[\heavyrulewidth] \midrule[\heavyrulewidth]
    \bf WMT'14 En-De    & Transformer-Base                & Transformer-Big \\\midrule
    Adam without warmup    & /         & /          \\
    Adam with warmup   & 26.81               & 28.15           \\
    AdaMod              & 27.22                           & 28.47           \\\bottomrule
    \end{tabular}
    \caption{BLEU score on Neural Machine Translation. ``/'' denotes divergence.}
    
    % \begin{tabular}{c|c|c|c}
    % \textbf{Method} & \textbf{IWSLT'14 De-En} & {\textbf{WMT'14 En-De}\\ (Transformer-base)}     &  {\textbf{WMT'14 En-De}\\ (Transformer-big)}\\
    % \hline
    % Adam with warmup    & 34.62 &26.81 &28.15 \\
    % AdaMod    & 35.14 & 27.22 & 28.47\\
    % \end{tabular}
    \label{tab:tab3}
\end{table}
%%%%%%%%%%%%%%%%%%%%%%%%%%%%%%%%%%%%%%%%%%%%%%%%%%%%%%%%%%%%%%%%%%%%%%%%%%%%%%%%%%%%%

%%%%%%%%%%%%%%%%%%%%%%%%%%%%%%%%%%%%%%%%%%%%%%%%%%%%%%%%%%%%%%%%%%%%%%%%%%%%%%%%%%%%%
\begin{figure}[t]
    \centering
    \subcaptionbox{Adam with warmup   \label{fig:mt-robust1}}{\includegraphics[width=0.4\textwidth]{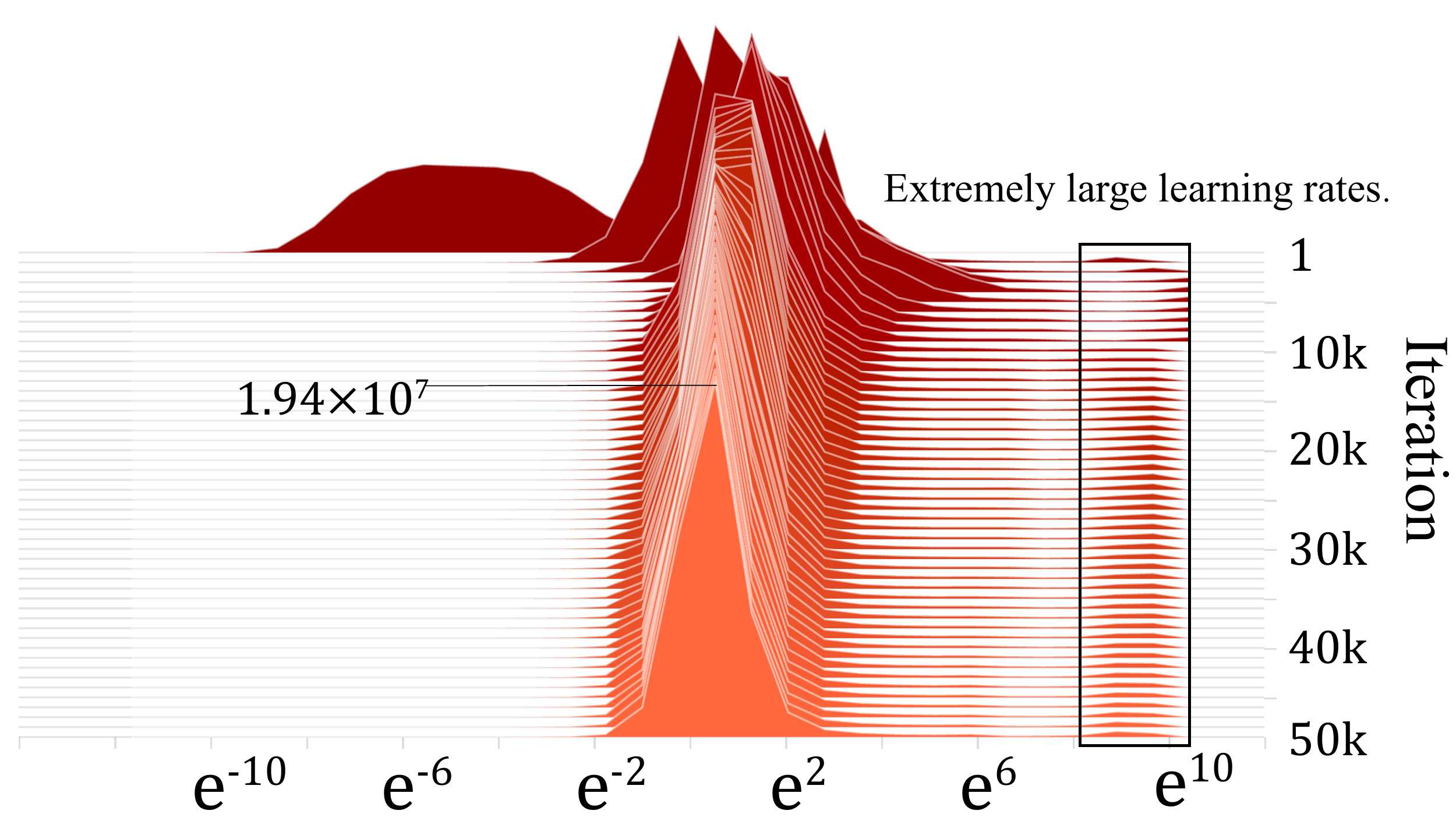}}
    
    \subcaptionbox{AdaMod   \label{fig:mt-robust2}}{\includegraphics[width=0.4\textwidth]{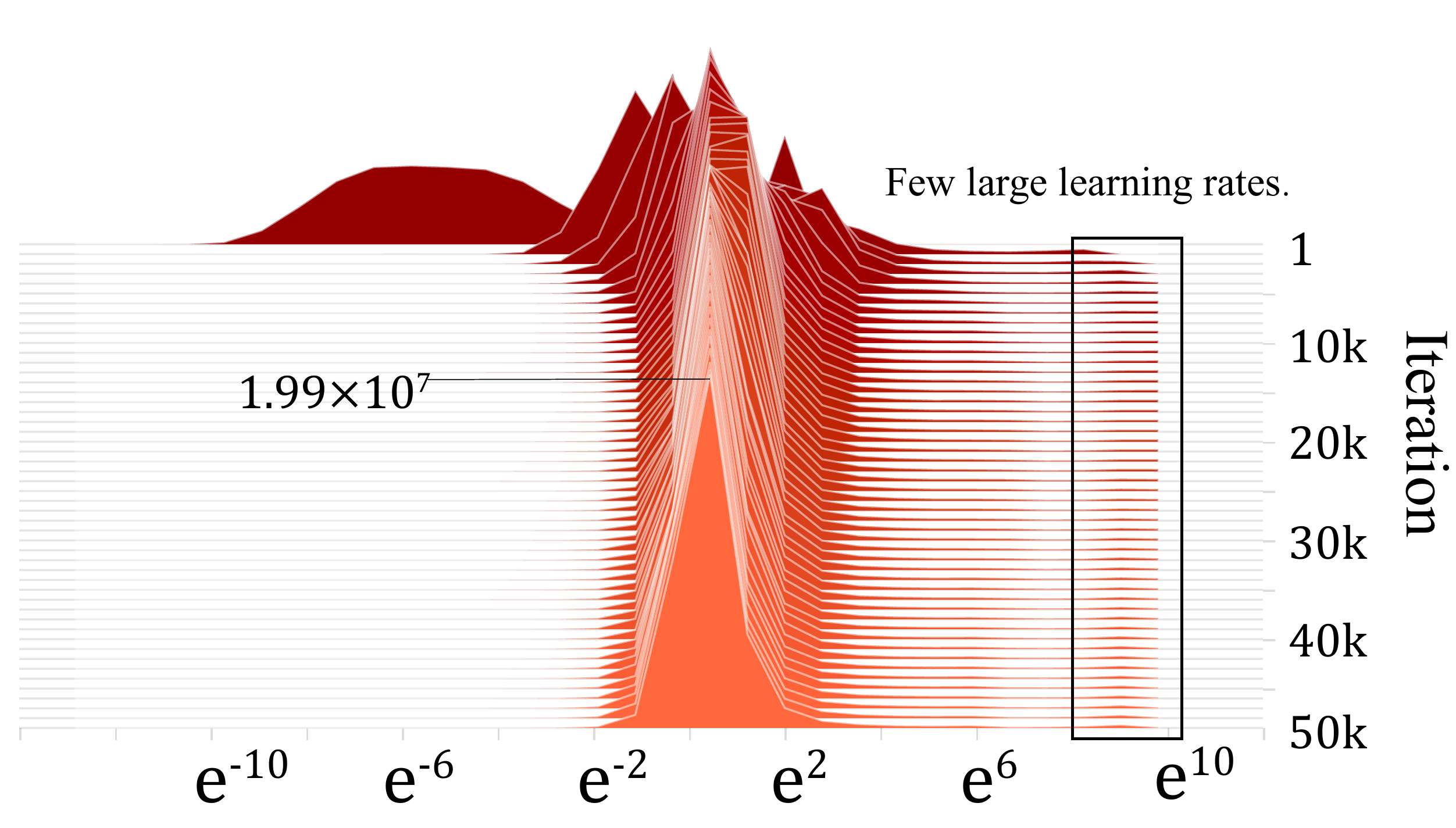}}

    \caption{The learning rate comparison of Transformers on the IWSLT'14 De-En. AdaMod properly restrains extremely large learning rates throughout the training process.}
    \label{fig:lr-comparison}
\end{figure}
%%%%%%%%%%%%%%%%%%%%%%%%%%%%%%%%%%%%%%%%%%%%%%%%%%%%%%%%%%%%%%%%%%%%%%%%%%%%%%%%%%%%%

%%%%%%%%%%%%%%%%%%%%%%%%%%%%%%%%%%%%%%%%%%%%%%%%%%%%%%%%%%%%%%%%%%%%%%%%%%%%%%%%%%%%%
\begin{figure*}[t]
    
    \subcaptionbox{ResNet-34 (Train) \label{fig:resnet1}}{\includegraphics[width=0.24\textwidth]{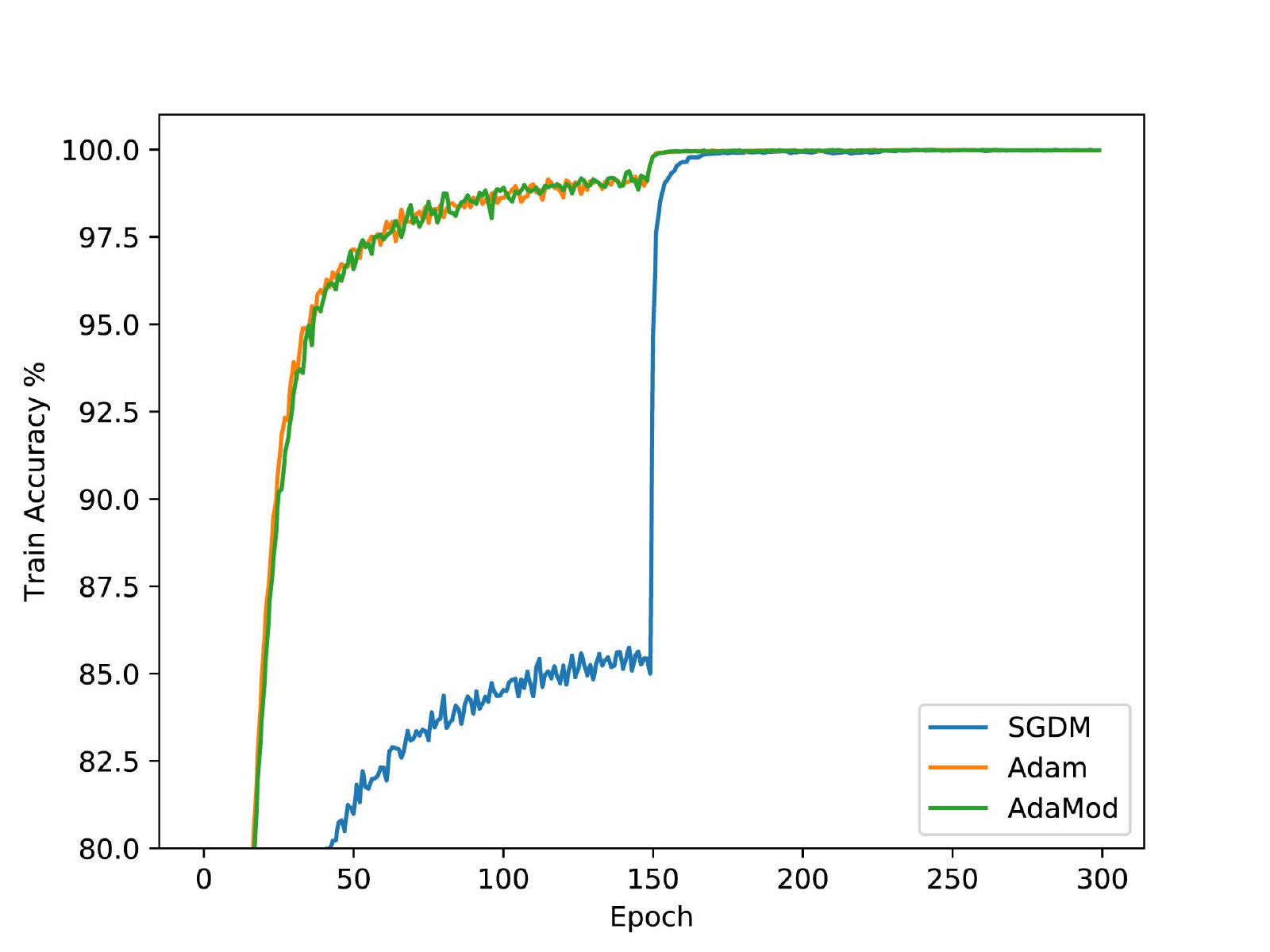}}
    \hfil
    \subcaptionbox{ResNet-34 (Test) \label{fig:resnet2}}{\includegraphics[width=0.24\textwidth]{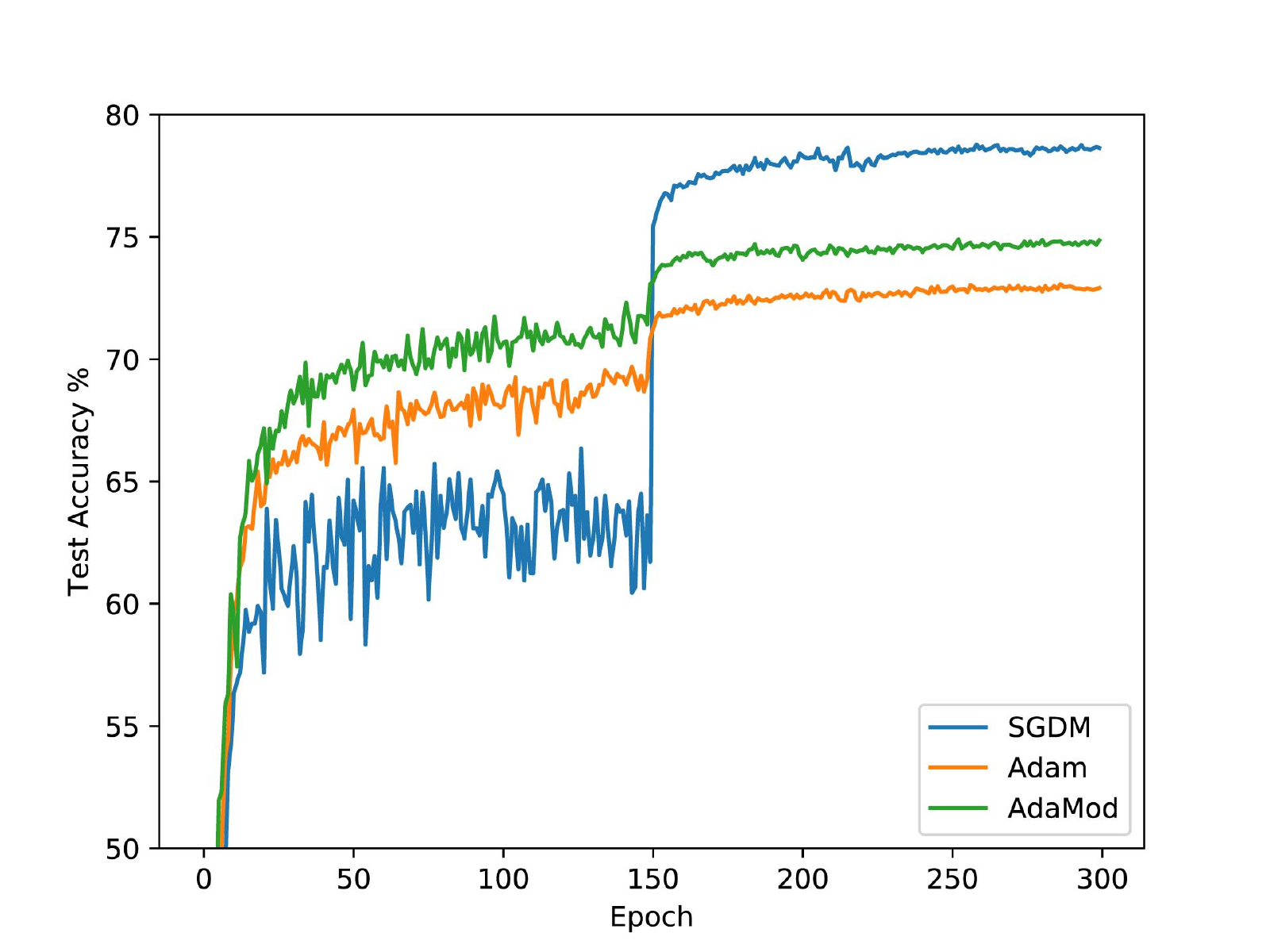}}
    \hfil
    \hfil
    \subcaptionbox{DenseNet-121 (Train) \label{fig:densenet1}}{\includegraphics[width=0.24\textwidth]{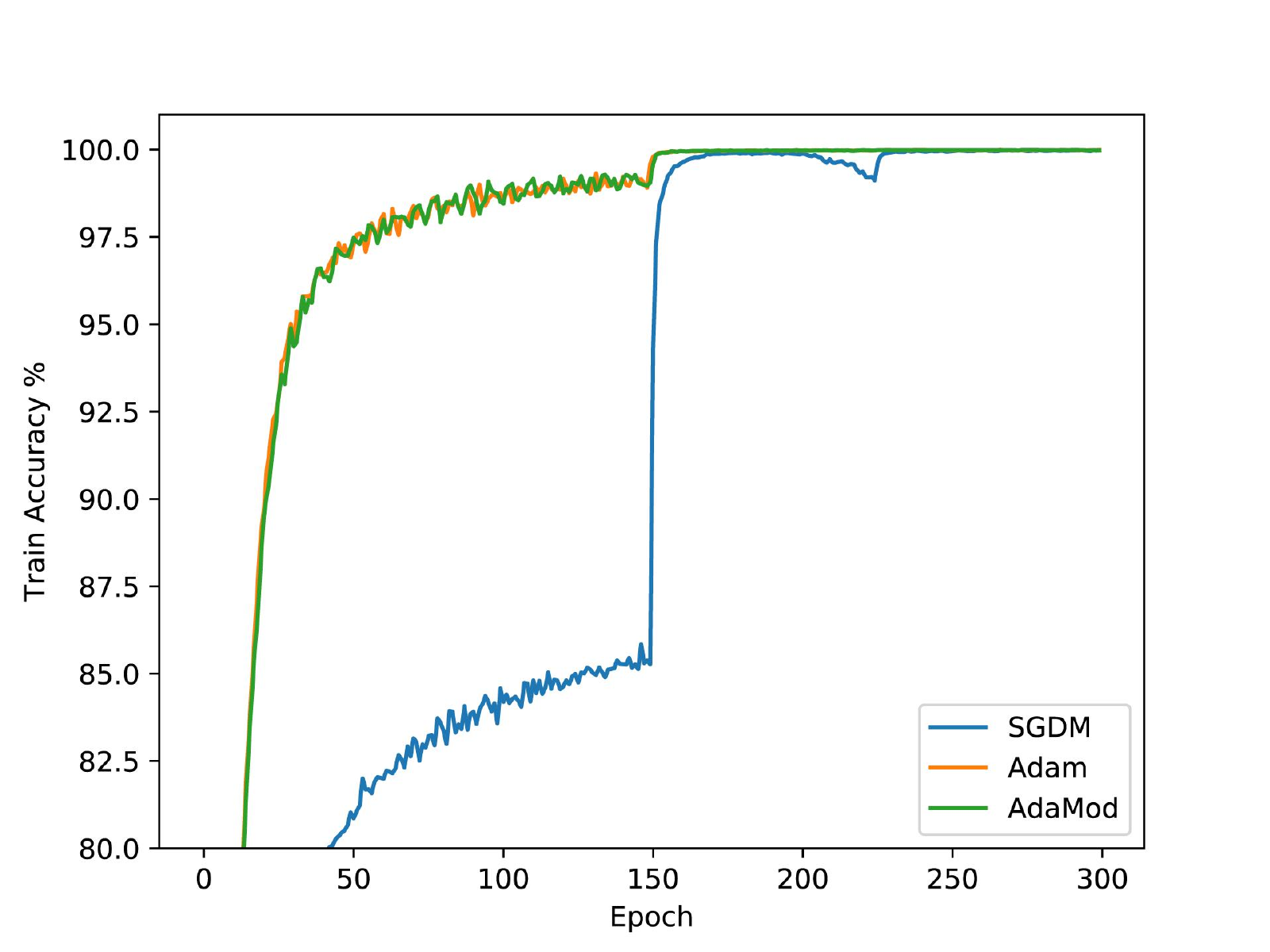}}
    \hfil
    \subcaptionbox{DenseNet-121 (Test) \label{fig:densenet2}}{\includegraphics[width=0.24\textwidth]{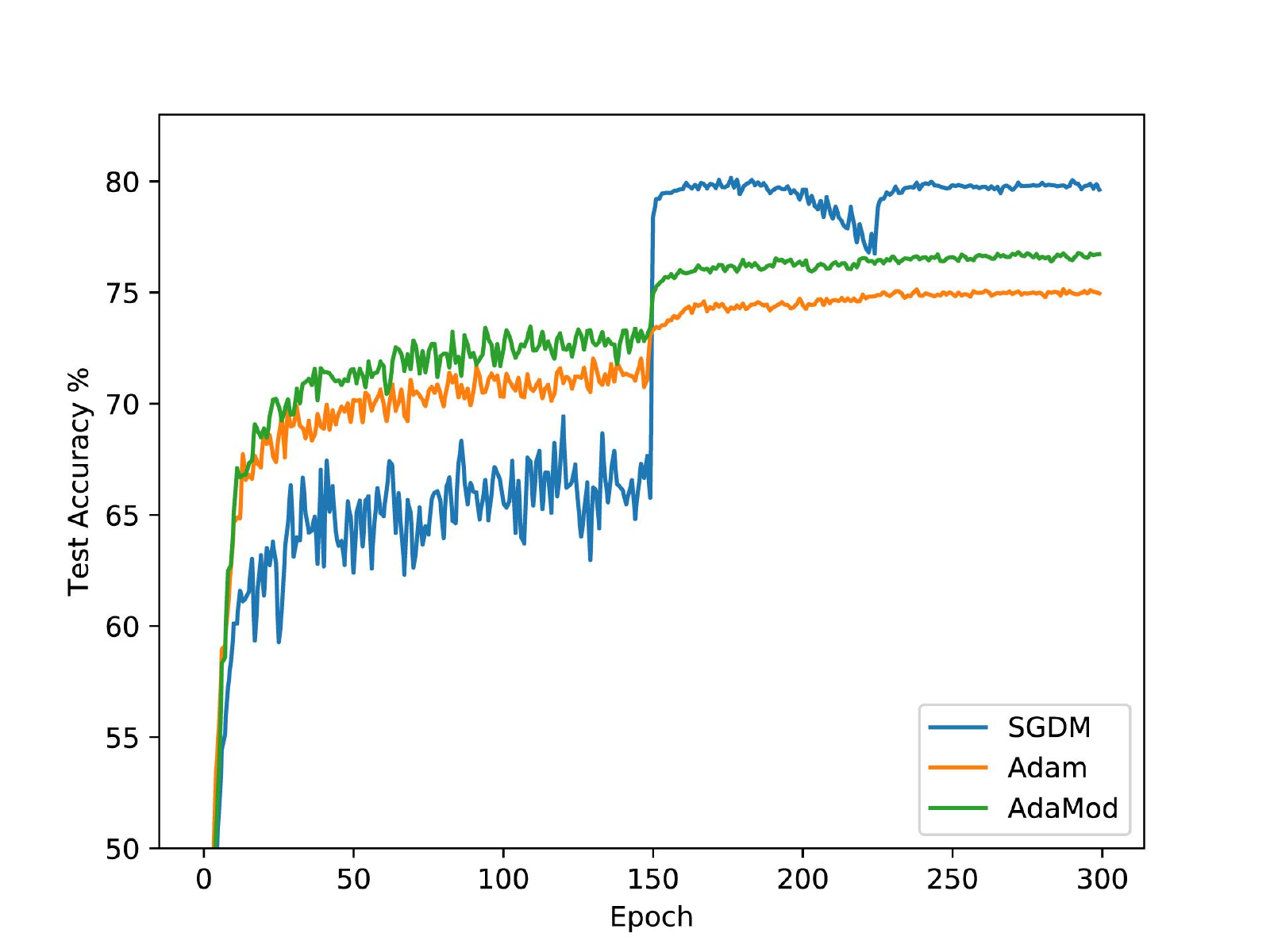}}
  
    \caption{Training and test accuracy for ResNet-34 and DenseNet-121 on CIFAR-100. AdaMod can achieve better accuracy both for ResNet and DenseNet on CIFAR-100 compared to Adam.}
    \label{fig:cifar100}
\end{figure*}
%%%%%%%%%%%%%%%%%%%%%%%%%%%%%%%%%%%%%%%%%%%%%%%%%%%%%%%%%%%%%%%%%%%%%%%%%%%%%%%%%%%%%

%%%%%%%%%%%%%%%%%%%%%%%%%%%%%%%%%%%%%%%%%%%%%%%%%%%%%%%%%%%%%%%%%%%%%%%%%%%%%%%%%%%%%
\begin{figure*}[t]

    \subcaptionbox{ResNet-34 (Train) \label{fig:resnet-train-triple9}}{
        \includegraphics[width=0.23\textwidth]{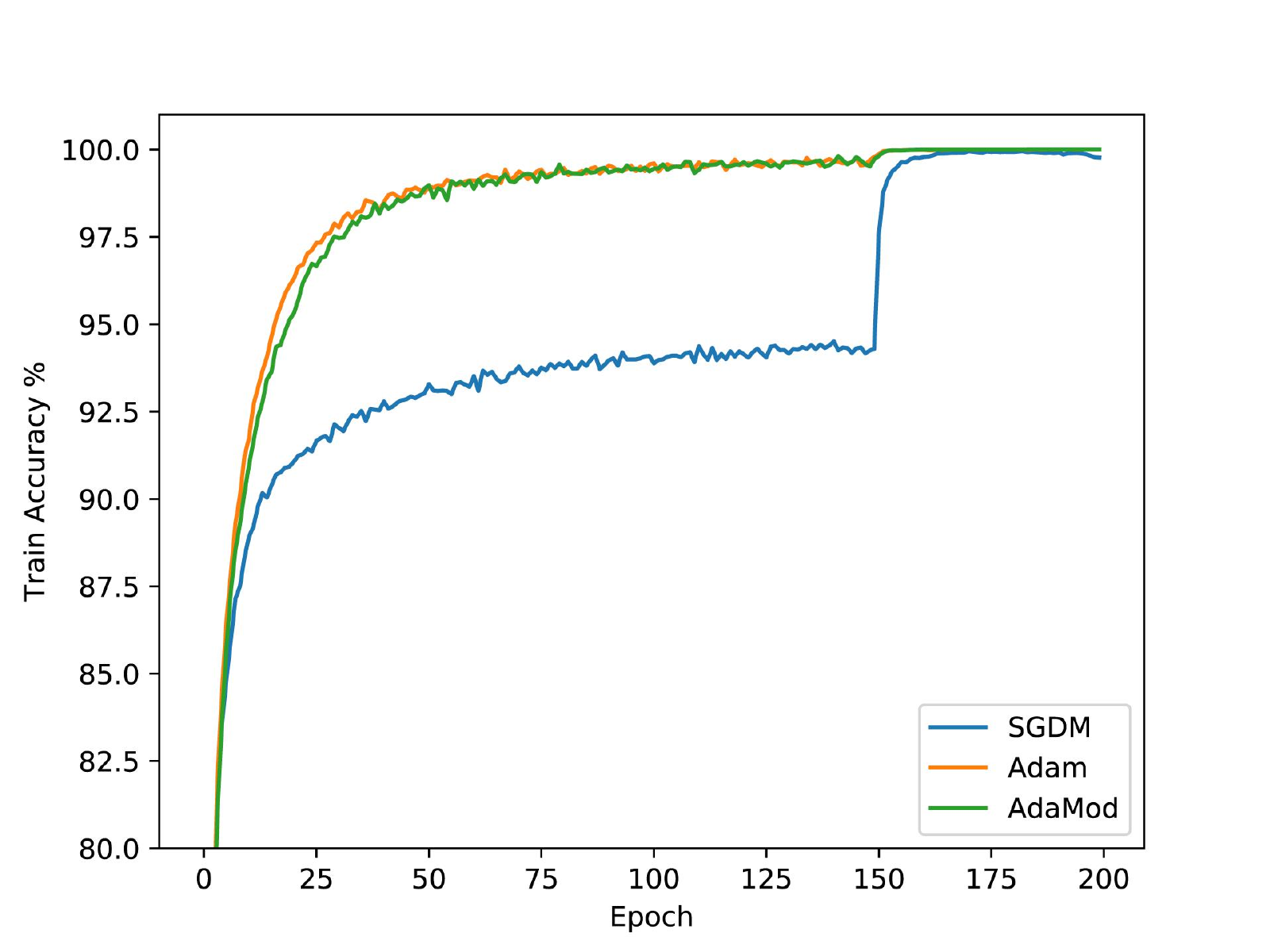}
    }
    \hfil
    \subcaptionbox{ResNet-34 (Test) \label{fig:resnet-test-triple9}}{
        \includegraphics[width=0.23\textwidth]{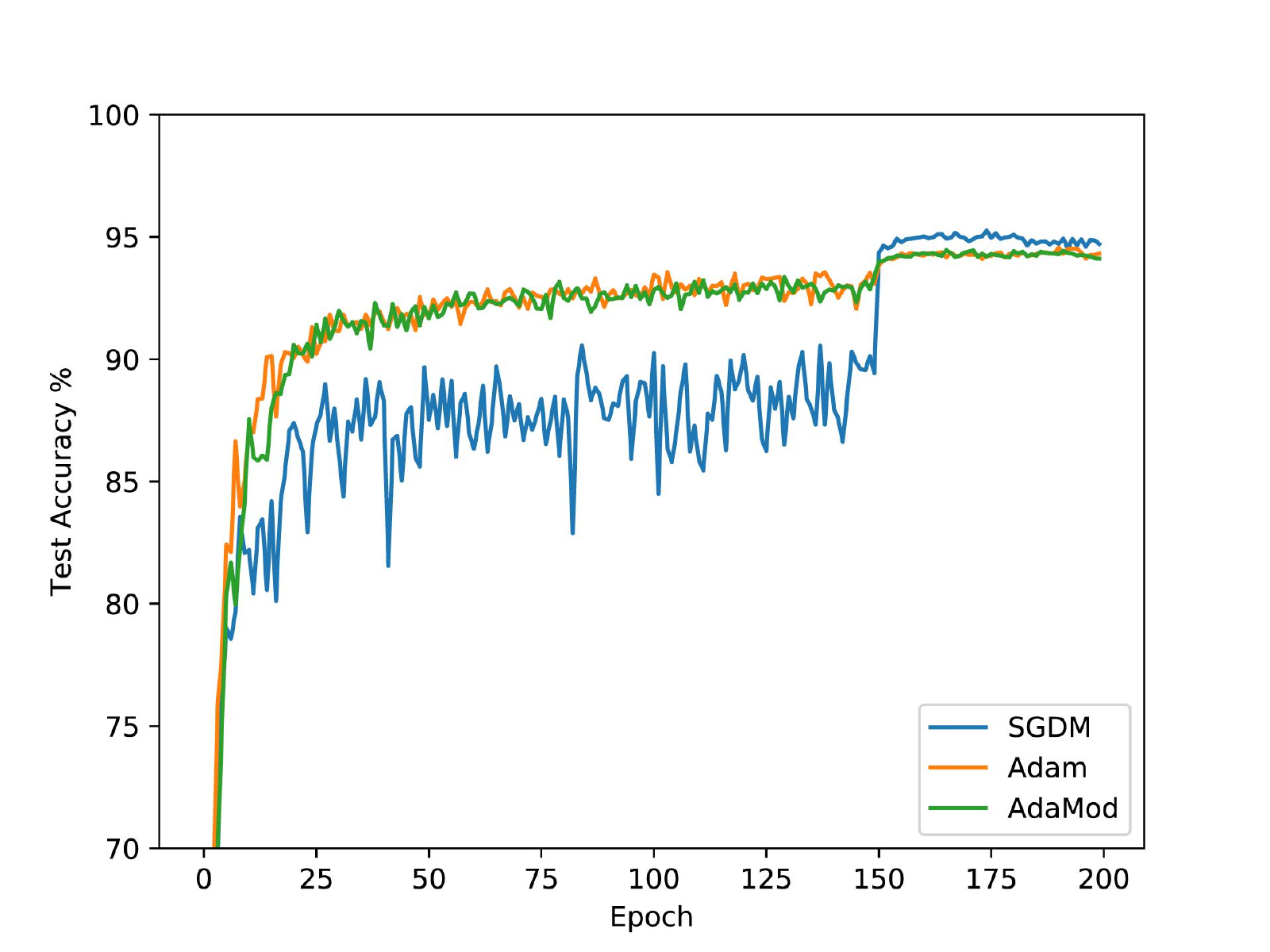}
    }
    \hfil
    \hfil
    \subcaptionbox{DenseNet-121 (Train) \label{fig:densenet-train-triple9}}{
        \includegraphics[width=0.23\textwidth]{cifar10-densenet-train_acc-min.pdf}
    }
    \hfil
    \subcaptionbox{DenseNet-121 (Test) \label{fig:densenet-test-triple9}}{
        \includegraphics[width=0.23\textwidth]{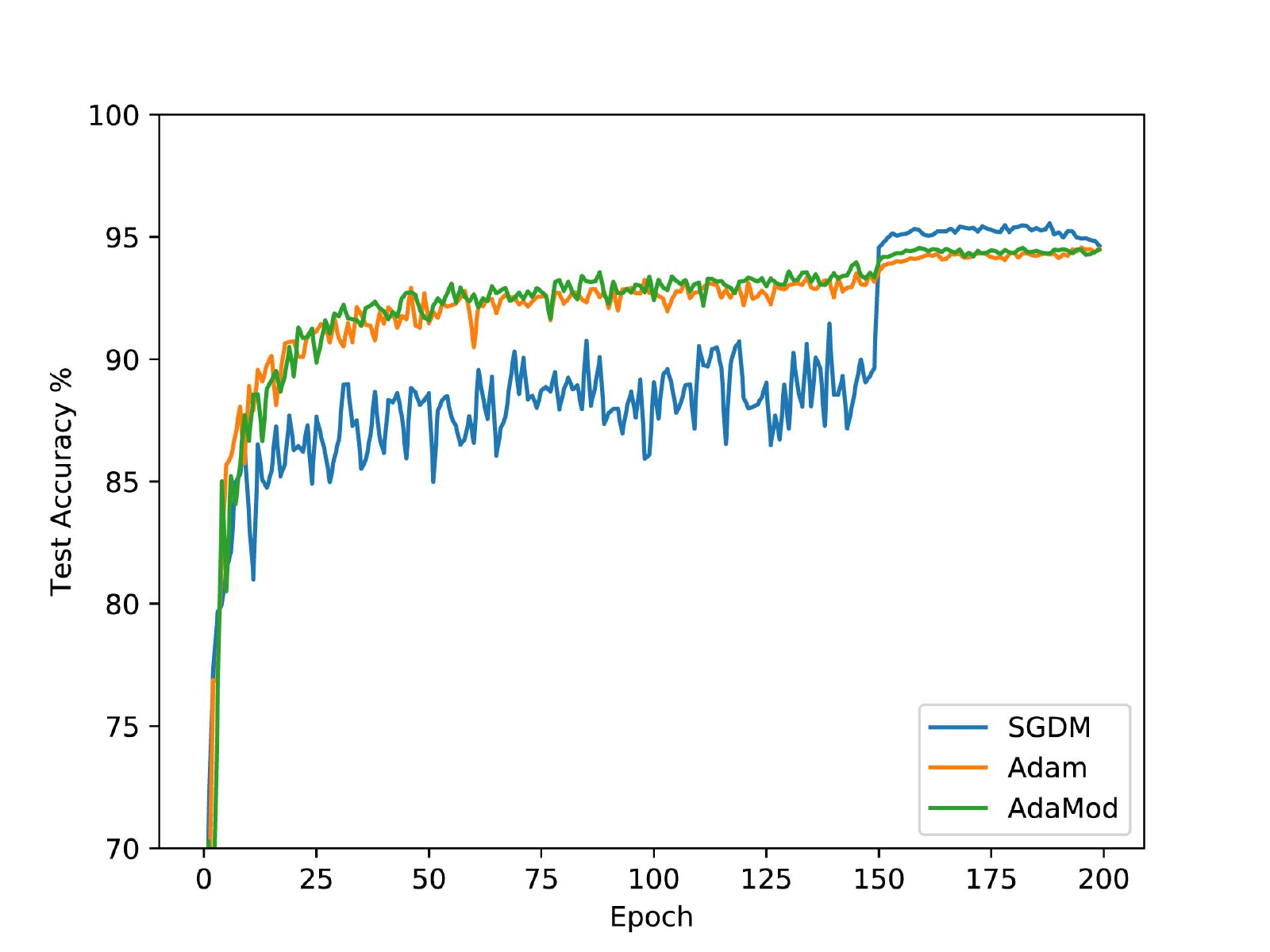}
    }

    \caption{Training and test accuracy for ResNet-34 and DenseNet-121 on CIFAR-10. AdaMod can achieve matched or better accuracy both for ResNet and DenseNet on CIFAR-10 compared to Adam.}
    \label{fig:cifar10}
\end{figure*}
%%%%%%%%%%%%%%%%%%%%%%%%%%%%%%%%%%%%%%%%%%%%%%%%%%%%%%%%%%%%%%%%%%%%%%%%%%%%%%%%%%%%%

%%%%%%%%%%%%%%%%%%%%%%%%%%%%%%%%%%%%%%%%%%%%%%%%%%%%%%%%%%%%%%%%%%%%%%%%%%%%%%%%%%%%%
\begin{figure*}[t]
    \hfil
    \subcaptionbox{SGDM with different $\alpha$    \label{fig:robust1}}{\includegraphics[width=0.3\textwidth]{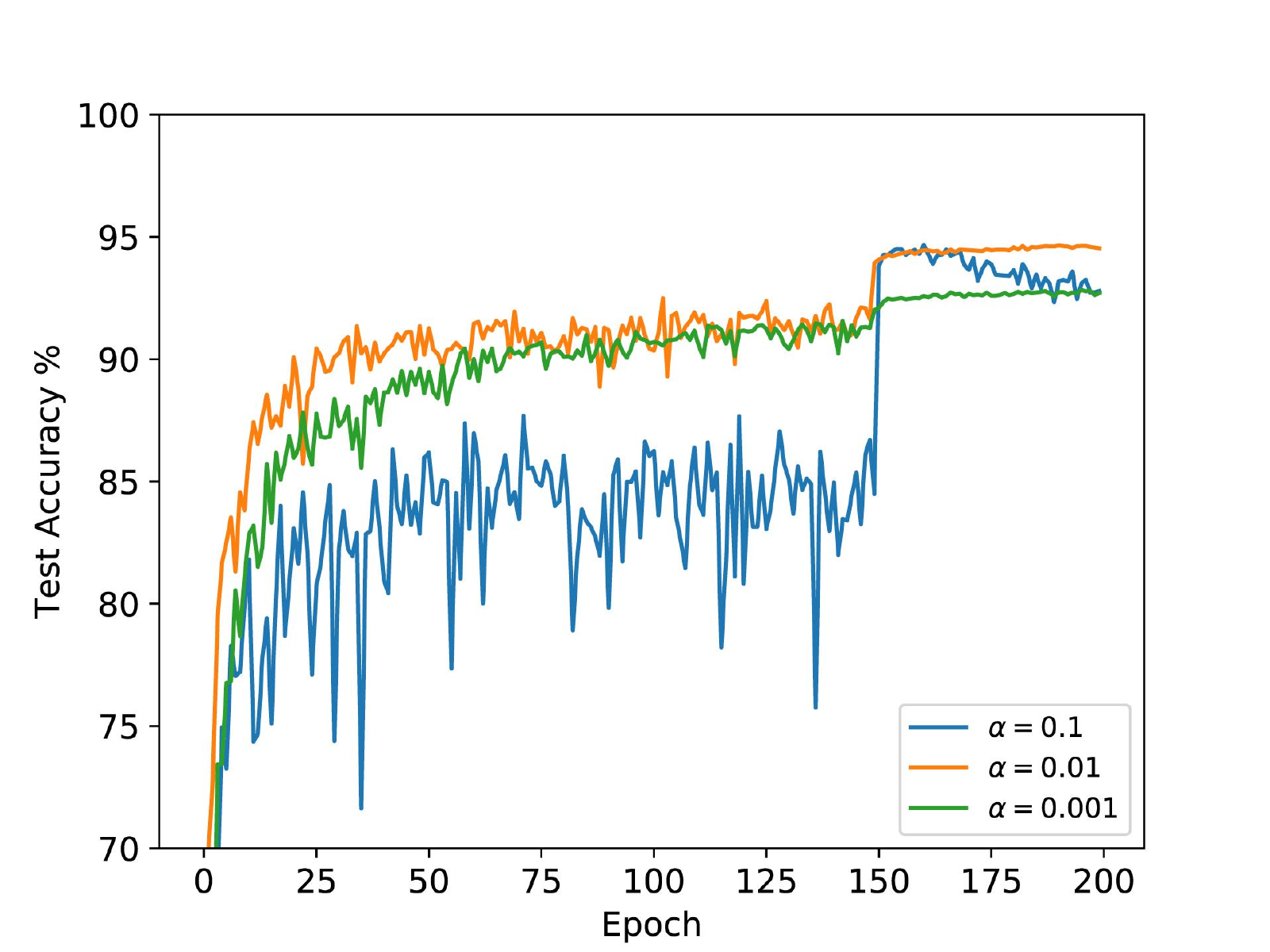}}
    \hfil
    \subcaptionbox{Adam with different $\alpha$    \label{fig:robust2}}{\includegraphics[width=0.3\textwidth]{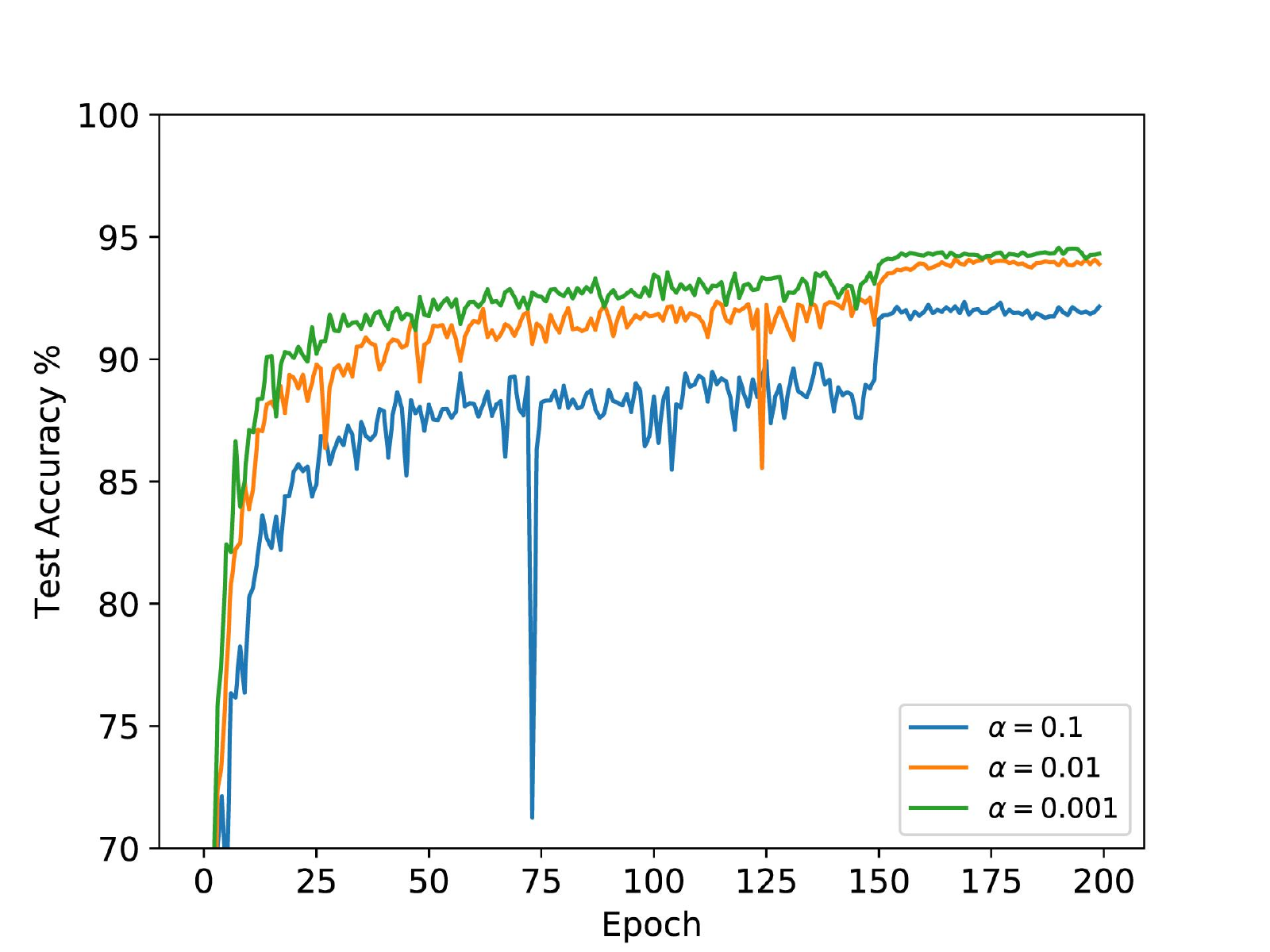}}
    \hfil
    \subcaptionbox{AdaMod with different $\alpha$    \label{fig:robust3}}{\includegraphics[width=0.3\textwidth]{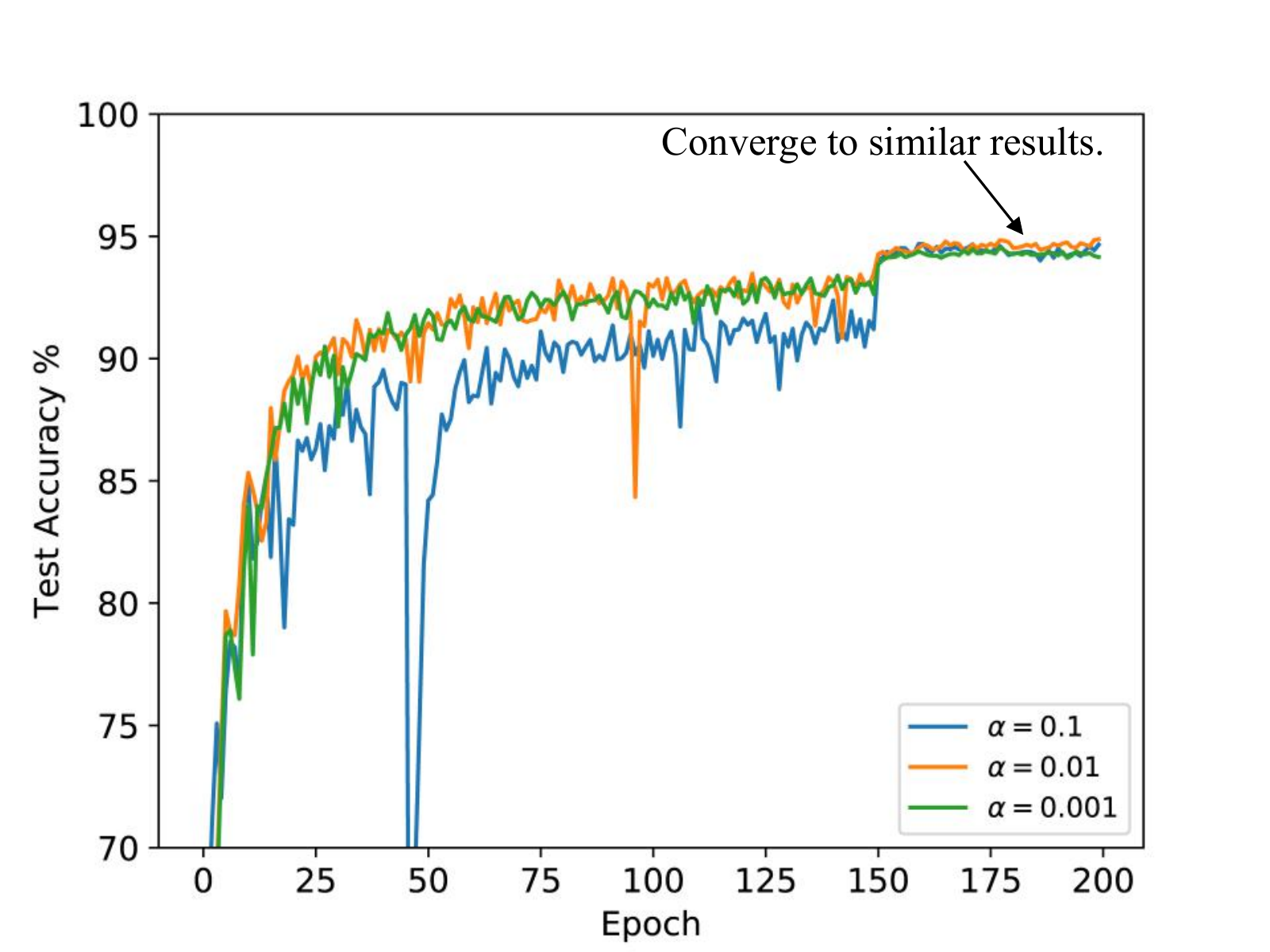}}
    \hfil
    
    \caption{Test accuracy of SGDM, Adam and AdaMod with different $\alpha$ using ResNet-34 on CIFAR-10. AdaMod more likely converges to similar results when $\alpha$ is different, which improves the robustness of model training.}
    \label{fig:resnet-different-lr}
\end{figure*}
%%%%%%%%%%%%%%%%%%%%%%%%%%%%%%%%%%%%%%%%%%%%%%%%%%%%%%%%%%%%%%%%%%%%%%%%%%%%%%%%%%%%%

\subsection{Neural Machine Translation}

Machine translation is one of the most important applications in natural language processing \citep{vaswani2017attention}. To evaluate the effectiveness of AdaMod, we train transformer-based models on two widely used datasets: IWSLT'14 De-En and WMT'14 En-De. 

Our experiments are based on the vanilla Transformers \citep{vaswani2017attention} implementation from the \textit{fairseq} open library \citep{DBLP:conf/naacl/OttEBFGNGA19}. Due to the limited size of the IWSLT'14 dataset, we use a relatively small model in training. The size of embeddings and hidden states is set to 512 and the number of heads in multi-head attention is set to 4. For WMT'14, we train the transformer base version and the big version respectively. Both of the two models consist of a 6-layer encoder and a 6-layer decoder. The size of the embedding is set to 512 for the base model and 1024 for the big. We maintain the hyper-parameter settings as the original paper (i.e. $\beta_{1}=0.9$, $\beta_{2}=0.98$, $\epsilon = 1e-9$). We use a linear warmup for Adam in the first 4000 updates but not for AdaMod. For IWSLT'14, the dropout rate is set as 0.3, weight decay as 1e-4 and maximum tokens per batch as 4000. As for WMT'14, we set maximum tokens as 3584.

\subsubsection{Performance Comparison} 
We use BLEU \citep{papineni2002bleu} as the metric to evaluate the performance and report results in Table \ref{tab:tab3}. As discussed above, Adam licenses to the warmup learning rate scheme when training Transfomer-based models to avoid the non-convergence problem. But for AdaMod, it can train these models without the warmup setting and achieve significantly higher BLEU scores on both two datasets. Moreover, training loss curves are shown in Figure \ref{fig:transformer-loss}. It shows that AdaMod achieves faster convergence against Adam throughout the whole training process. In other words, AdaMod obtains considerable improvement over Adam on neural machine translation tasks by fixing the non-convergence issue.     

\subsubsection{Learning Rates Comparison} 
In order to verify the amelioration of adaptive learning rates of our method, we further compare the learning rates histogram of Transformers on the IWSLT'14 De-En between Adam and AdaMod, as shown in Figure \ref{fig:lr-comparison}, where the X-axis is original value in the log scale, and Y-axis is iteration steps and the height stands for frequency. Intuitively, AdaMod smooths out the unexpected large learning rates in the whole training process and brings consistent improvements. Specifically, in the early stage, AdaMod stabilizes the learning rates so it can be independent of warmup assistance. This reduces the hyperparameters of training and saves a lot of tuning time. In the middle and late terms, AdaMod keeps this good advantage and gets better generalization performance. 

%%%%%%%%%%%%%%%%%%%%%%%%%%%%%%%%%%%%%%%%%%%%%%%%%%%%%%%%%%%%%%%%%%%%%%%%%%%%%%%%%%%%%
\begin{table}[t]
    \centering
    \caption{Test accuracy for ResNet-34 and DenseNet-121 on CIFAR-100. Report for $Median\ (Mean \pm Std)$.}
    \begin{tabular}{@{}lcc@{}}
    \toprule
    \textbf{CIFAR-100}     & \textbf{ResNet-34}  & \textbf{DenseNet-121}\\
    \midrule
    SGDM    & $78.50\ (78.48\pm0.23)$ & $80.00\ (79.53\pm0.94)$ \\
    Adam     & $73.81\ (73.36\pm0.64)$ & $74.95\ (75.23\pm0.42)$ \\
    AdaMod   & $74.86\ (74.83\pm0.09)$ & $77.28\ (77.12\pm0.29)$ \\
    \bottomrule
    \end{tabular}
    \label{tab:tab6}
\end{table}
%%%%%%%%%%%%%%%%%%%%%%%%%%%%%%%%%%%%%%%%%%%%%%%%%%%%%%%%%%%%%%%%%%%%%%%%%%%%%%%%%%%%%

%%%%%%%%%%%%%%%%%%%%%%%%%%%%%%%%%%%%%%%%%%%%%%%%%%%%%%%%%%%%%%%%%%%%%%%%%%%%%%%%%%%%%
\begin{table}[t]
    \centering
    \caption{Test accuracy for ResNet-34 and DenseNet-121 on CIFAR-10. Report for $Median\ (Mean \pm Std)$.}
    \begin{tabular}{@{}lcc@{}}
    \toprule
    \textbf{CIFAR-10}     & \textbf{ResNet-34}  & \textbf{DenseNet-121}\\
    \midrule
    SGDM    & $94.48\ (94.52\pm0.14)$ & $94.47\ (94.48\pm0.12)$ \\
    Adam     & $94.31\ (94.40\pm0.15)$ & $94.52\ (94.47\pm0.15)$ \\
    AdaMod   & $94.30\ (94.29\pm0.14)$ & $94.72\ (94.68\pm0.08)$ \\
    \bottomrule
    \end{tabular}
    \label{tab:tab5}
\end{table}
%%%%%%%%%%%%%%%%%%%%%%%%%%%%%%%%%%%%%%%%%%%%%%%%%%%%%%%%%%%%%%%%%%%%%%%%%%%%%%%%%%%%%

\subsection{Image Classification}
We consider the task of image classification on CIFAR-10 and CIFAR-100 datasets. For CIFAR-10 experiments, we train the model with 200 epochs on ResNet-34 \citep{he2016deep} and DenseNet-121 \citep{huang2017densely} respectively with batches of 128 images and decay the learning rates by 10 at the 150\textit{th} epoch. Similarly, for CIFAR-100, we employ 300 epochs on the two models with the same batch size but reduce the learning rates by 10 both at the 150\textit{th} and the 225\textit{th} epoch. For Adam and AdaMod, we set $\beta_{1}=0.9$, $\beta_{2}=0.999$. For SGD, we configure the momentum factor as 0.9. We apply a weight decay of 5e-4 to all the methods. In addition, we conduct experiments using 3 random seeds and report their key features, i.e. $Median\  (Mean \pm Std)$ . Our results are summarized in Table \ref{tab:tab6} and Table \ref{tab:tab5}.

\subsubsection{ResNet}
The accuracy curves are shown in Figure \ref{fig:cifar100} and Figure \ref{fig:cifar10}. We can see that AdaMod outperforms Adam almost in both two datasets especially on CIFAR-100. Although the upper bounds of learning rates limit the speed of AdaMod in the early epochs, it can also catch up with Adam in the mid-term and achieves best training accuracy after learning rates are decayed. More importantly, our method gets both faster convergence and better performance than Adam on the test set, which verifies the consistent improvement on stabilizing learning rates of entire training process. Note that on these two datasets, SGDM usually behaves better than adaptive methods \citep{wilson2017marginal,keskar2017improving,DBLP:conf/iclr/LuoXLS19}. Despite AdaMod fails to compete with SGDM in the test accuracy, it shows better training performance. 

\subsubsection{DenseNet}
The accuracy curves for this experiment are summarized in Figure \ref{fig:cifar100} and Figure \ref{fig:cifar10}. As we expect, the overall performance of AdaMod on DenseNet-121 is even better than on ResNet, and the improvement of AdaMod relative to Adam becomes more significant, which is enhanced with more than 2\% in the test accuracy on CIFAR-100. And on CIFAR-10, AdaMod outperforms SGDM and win the top performance. These serve as evidences that AdaMod gains more benefits with the enrichment of model's complexity. To sum up, AdaMod can achieve matched or better accuracy for both ResNet and DenseNet on CIFAR-10/CIFAR-100 datasets. In other words, even there is no non-convergence problem in the early training stage (e.g. without warmup assistance), it is obviously beneficial to smooth and stabilize the adaptive learning rates throughout the training.

%%%%%%%%%%%%%%%%%%%%%%%%%%%%%%%%%%%%%%%%%%%%%%%%%%%%%%%%%%%%%%%%%%%%%%%%%%%%%%%%%%%%%
\begin{table}[t]
    \centering
    \caption{Test perplexity on Language Modeling. Report for $Median\ (Mean \pm Std)$.}
    \begin{tabular}{@{}lc@{}}
    \toprule
    \textbf{Penn Treebank}     & \textbf{LSTM} \\
    \midrule
    Adam     & $71.08\ (70.95\pm0.27)$ \\
    AdaMod   & $70.78\ (70.76\pm0.10)$ \\
    \bottomrule
    \end{tabular}
    \label{tab:tab2}
\end{table}
%%%%%%%%%%%%%%%%%%%%%%%%%%%%%%%%%%%%%%%%%%%%%%%%%%%%%%%%%%%%%%%%%%%%%%%%%%%%%%%%%%%%%

%%%%%%%%%%%%%%%%%%%%%%%%%%%%%%%%%%%%%%%%%%%%%%%%%%%%%%%%%%%%%%%%%%%%%%%%%%%%%%%%%%%%%
\begin{figure}[ht]
    \centering
    \includegraphics[width=0.4\textwidth]{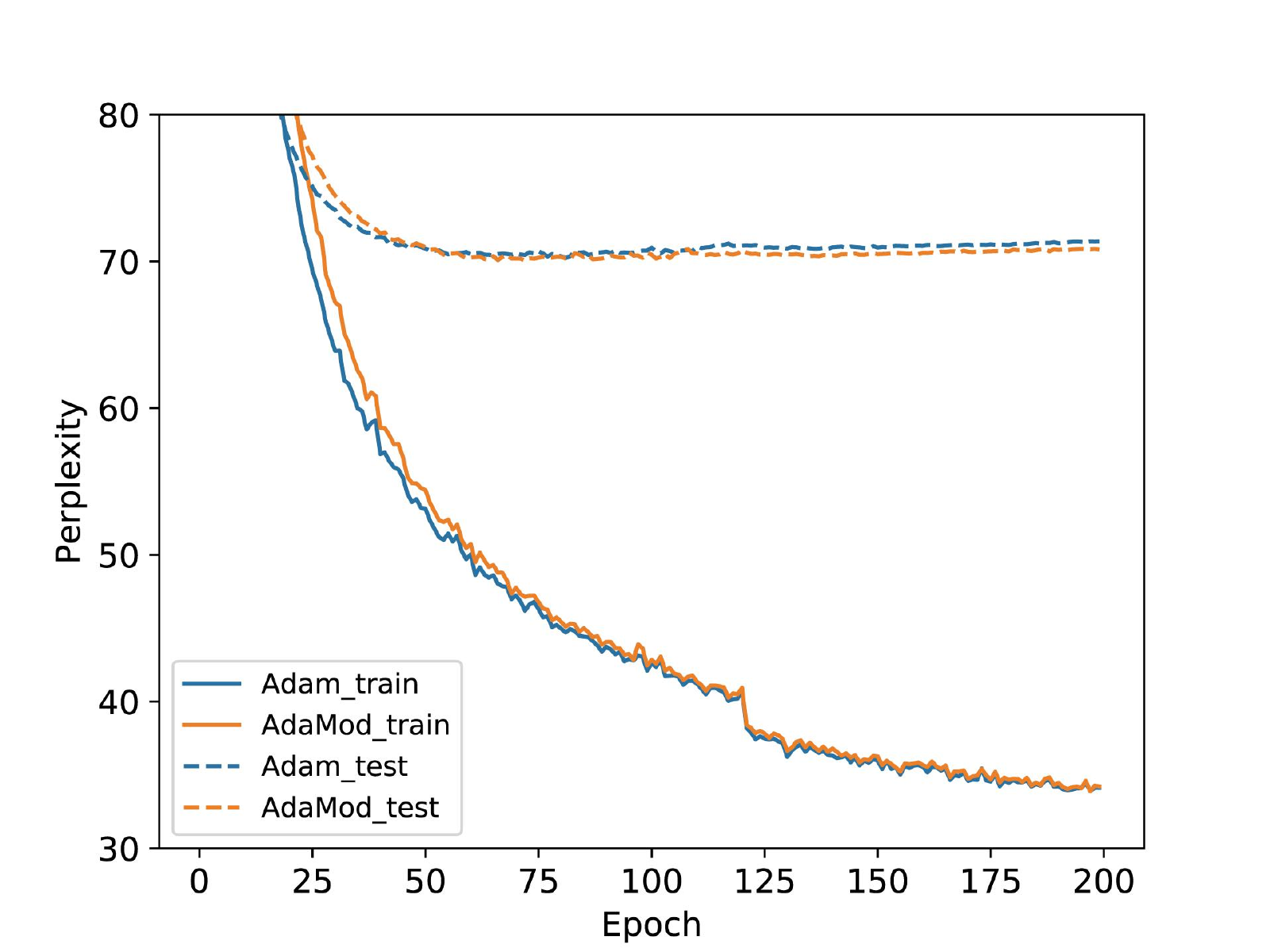}
    \caption{Training and test perplexity for 3-layer LSTM on Penn Treebank.}
    \label{fig:fig3}
\end{figure}
%%%%%%%%%%%%%%%%%%%%%%%%%%%%%%%%%%%%%%%%%%%%%%%%%%%%%%%%%%%%%%%%%%%%%%%%%%%%%%%%%%%%%

\subsection{Language Modeling}

We also conduct an experiment on the language modeling task. Specifically, we train a 3-layer LSTM network with 3450 hidden states \citep{hochreiter1997long} on the Penn Treebank dataset, running for 200 epochs and reduce the learning rates by 10 at the 120\textit{th} epoch. Following the setup of \citet{DBLP:conf/iclr/MerityKS18}'s work, we set batch size as 20 and $\beta_{1}=0.9$, $\beta_{2}=0.999$. We adopt their public code and run experiments for 3 random seeds in this study. The perplexities are summarized in Table \ref{tab:tab2}. It is worthy noting that we do not exert fine-tuning and continuous cache pointer augmentation \citep{DBLP:conf/iclr/MerityKS18} on these results.

Perplexity curves are displayed in Figure \ref{fig:fig3}. It shows that AdaMod lags behind Adam in the early stage, but AdaMod gradually outperforms Adam with the increase of steps. The experiments demonstrate the versatility of AdaMod on different tasks, although it is slightly better than Adam in terms of training speed and generalization performance. 

\subsection{Analysis}
\paragraph{Robustness to different learning rates}
To investigate the robustness of AdaMod, we conduct experiments with the ResNet-34 model on the CIFAR-10 dataset. We test SGDM, Adam and AdaMod with different $\alpha$ (i.e. initial learning rate), which is chosen in \{0.1, 0.01, 0.001\} and $\beta_{3} = 0.9999$ for AdaMod. The results are displayed in Figure \ref{fig:resnet-different-lr}. It is observed that SGDM and Adam are sensitive to the hyperparameter. Especially as $\alpha$ becomes larger, the performance gap among the different learning rates becomes more noticeable. The phenomenon also confirms the previous results that adopting a suitable learning rate is vital for SGDM, as both the small learning rate ($\alpha=0.001$) and the large learning rate ($\alpha=0.1$) lead to significantly worse results. Our results also show that Adam is more friendly for smaller learning rates and more or less stable, e.g. $\alpha=0.001$ is slightly better than $\alpha=0.01$, while it performs much less stable when alpha is too large, e.g., $\alpha=0.1$ due to the extremely large learning rates. By contrast, AdaMod has almost identical final test accuracy for those $\alpha$ within a broad range, which demonstrates the robustness of AdaMod with respect to initial learning rates and supports our motivation that dealing with extremely large learning rates in Adam is very beneficial.

\paragraph{Robustness to different $\boldsymbol{\beta_{3}}$}

%%%%%%%%%%%%%%%%%%%%%%%%%%%%%%%%%%%%%%%%%%%%%%%%%%%%%%%%%%%%%%%%%%%%%%%%%%%%%%%%%%%%%
\begin{figure}[t]
    \centering
    \includegraphics[width=0.4\textwidth]{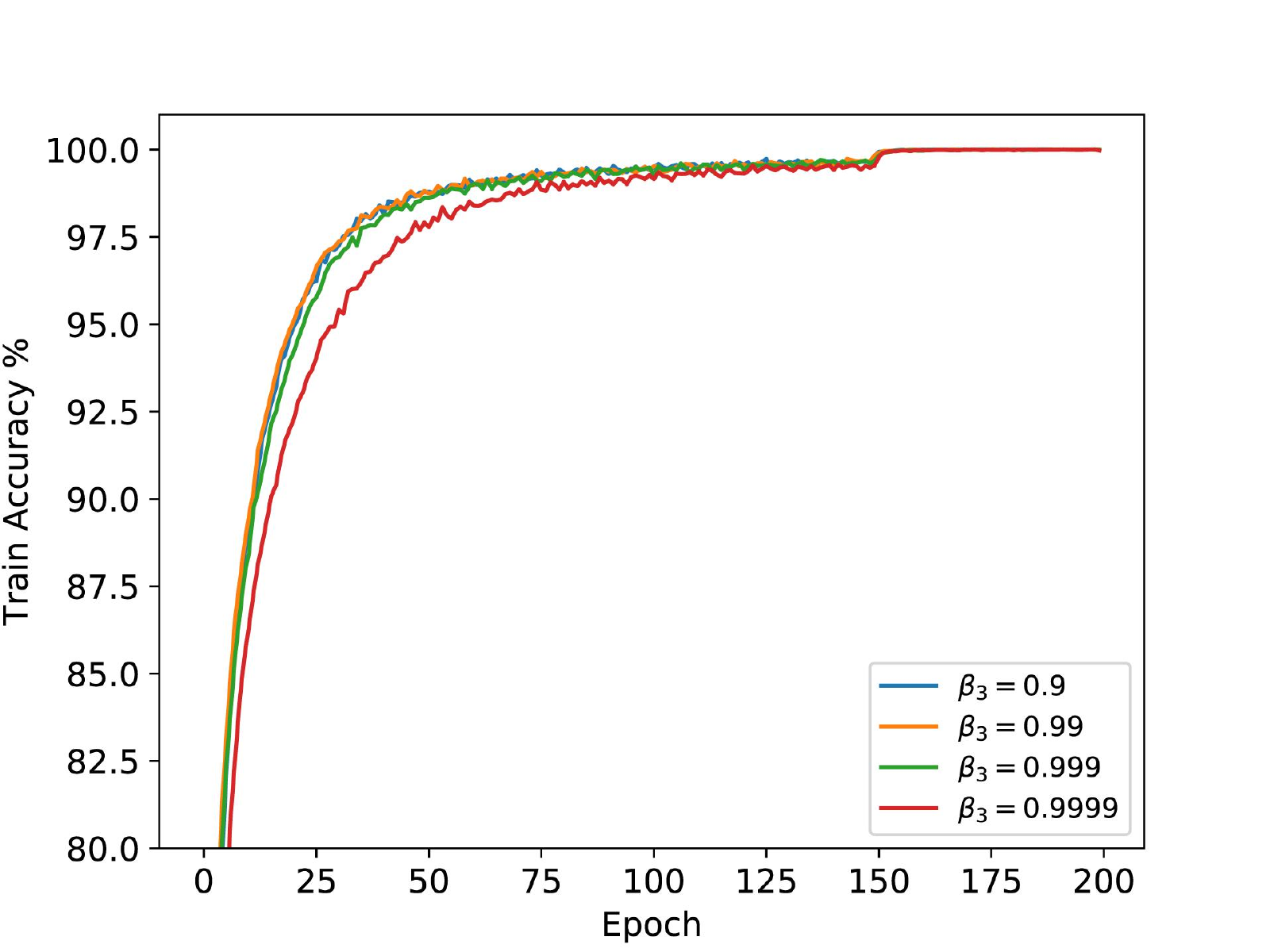}
    \caption{Training accuracy of AdaMod with different \textbf{$\beta_{3}$} using ResNet-34 on CIFAR10.}
    \label{fig:sensitive1}
\end{figure}
%%%%%%%%%%%%%%%%%%%%%%%%%%%%%%%%%%%%%%%%%%%%%%%%%%%%%%%%%%%%%%%%%%%%%%%%%%%%%%%%%%%%%

\begin{figure}[t]
    \centering
    \includegraphics[width=0.4\textwidth]{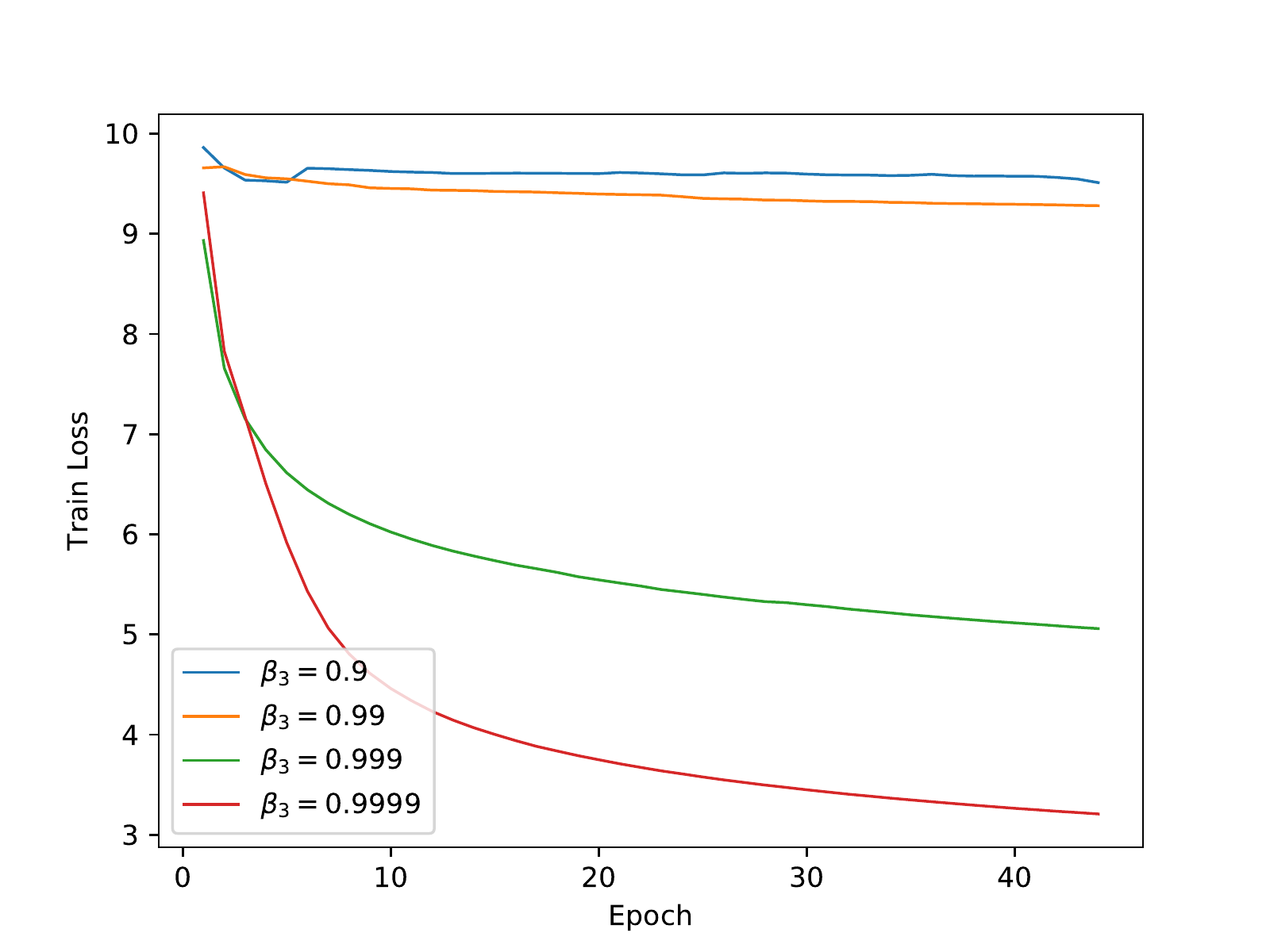}
    \caption{Training loss of AdaMod with different \textbf{$\beta_{3}$} using Transformer-small on IWSLT'14.}
    \label{fig:sensitive2}
\end{figure}

Furthermore, we investigate the impact of $\mathbf{\beta_{3}}$ and the results are displayed in Figure \ref{fig:sensitive1} and \ref{fig:sensitive2}.
We first test AdaMod with different $\beta_{3}$ with the ResNet-34 model, where $\beta_{3}$ are chosen in \{0.9,0.99,0.999,0.9999\} and $\alpha = 0.001$. We can see that for a specific $\alpha$, larger $\beta_{3}$ results in a lower convergence speed, but the performances with different $\beta_{3}$ are very close. It indicates that the convergence speed shows minor effect to the final results in most of the tasks. While for neural machine translation experiments, we test AdaMod with different $\beta_{3}$ with the Transformer-small model, where $\beta_{3}$ are chosen in \{0.9,0.99,0.999,0.9999\} and $\alpha = 0.0005$. It can be seen that when $\beta_{3}$ is small, the training loss converges to a poor result like Adam without warmup. As $\beta_{3}$ increases, the improvement of AdaMod over Adam is increasingly obvious, and achieves the best when $\beta_{3}=0.9999$. \textbf{Therefore, we recommend a $\boldsymbol{\beta_{3}}$ in \{0.999,0.9999\} as preferred for its usually behaving a strong performance across most models in practice}. That is, AdaMod can achieve a higher or matched performance to Adam even if without carefully fine-tuning.

In fact, $\beta_{3}$ controls the length of the gradient historical statistics used by the momental upper bound of the learning rate. In other words, a large $\beta_{3}$ endows learning rates with ``long-term memory''. As $\beta_{3}$ increases, this ``long-term memory'' becomes more dominant and the role of gradient historical statistics becomes more salient. For example, when $\beta_{3}=0.9999$, up to 10,000 steps of historical statistics will be taken into account. The benefit of this is that the momental upper bound of the learning rate fluctuates less and becomes more stable, thus greatly smoothing out the extremely large learning rates. 

%%%%%%%%%%%%%%%%%%%%%%%%%%%%%%%%%%%%%%%%%%%%%%%%%%%%%%%%%%%%%%%%%%%%%%%%%%%%%%%%%%%%%
%%%%%%%%%%%%%%%%%%%%%%%%%%%%%%%%%%%%%%%%%%%%%%%%%%%%%%%%%%%%%%%%%%%%%%%%%%%%%%%%%%%%%

\section{Future Work}
Although our method has improved in many aspects compared to Adam, there are still several problems to be explored. For example, the performance on many simple models still has a gap with SGDM, and how can we bridge the gap while maintaining our existing strengths? Also, we found that when $\beta_{3}$ is gradually increased within a certain range, the generalization performance of AdaMod tends to improve, yet with the cost of lowering the convergence speed. How can we tackle this trade-off relationship (e.g. design a proper scheduler to control it)? Besides, it is worth noting that AdaMod fixes the stability issue in optimization perspective rather than neural architectures. In such case, can we combine AdaMod with other orthogonal stabilization methods such as fixup initialization \citep{DBLP:conf/iclr/ZhangDM19} to achieve better performance? These deserve to be discussed.

\section{Conclusion}
In this paper, we study the warmup heuristic scheme used for adaptive optimization methods when training complex networks and identify the extremely large learning rates existing in the early training stage, which could hamper performance and lead to divergence. An empirical evidence is provided to support our hypothesis.

We design a concise strategy to constrain the learning rates of Adam to avoid the non-convergence issue. Our proposed algorithm, AdaMod, exerts adaptive upper bounds on individual learning rates to prevent them becoming undesirably larger than what the historical statistics suggest, leading to a better performance. Strong empirical results on many deep learning applications demonstrate the effectiveness of our proposed method especially on complex networks such as DenseNet and Transformer.

\section{Acknowledgments}
We are grateful to Liangchen Luo, Zhiyuan Zhang and Guangxiang Zhao for their helpful discussions. Xu Sun is the corresponding author of this paper.

% \appendix

% \section{Implementation Details}\label{app:app1}

\bibliography{aaai}
\bibliographystyle{aaai}

\end{document}